\documentclass[10pt,twocolumn,letterpaper]{article}

\usepackage[pagenumbers]{iccv} %

\usepackage[accsupp]{axessibility}
\usepackage{graphicx}
\usepackage{multirow,tabularx}
\usepackage[point]{fltpoint}
\usepackage{todonotes}
\usepackage{textcomp}
\usepackage{amsmath,amssymb}
\usepackage{ifthen}
\usepackage{color}
\usepackage{xcolor}
\usepackage{booktabs}
\usepackage{longtable}
\usepackage{footmisc}
\usepackage{enumitem}

\usepackage{microtype}
\usepackage{epsfig}
\usepackage{algorithm}
\usepackage{algpseudocode}

\DeclareGraphicsExtensions{.pdf,.jpg,.png}
\graphicspath{{./figures/}}

\newcolumntype{C}{X<{\centering}}
\newcolumntype{?}{!{\vrule width 1.0pt}}

\renewcommand\etal[1]{et al.~\cite{#1}}
\makeatletter
\DeclareRobustCommand\onedot{\futurelet\@let@token\@onedot}
\def\@onedot{\ifx\@let@token.\else.\null\fi\xspace}
\makeatother

\def\vs{\emph{vs}\onedot}

\newcommand\Caption[3][]{\caption[#2]{\label{#1}\textsc{#2}. \small#3}}

\renewcommand\sec[1]{Sec.~\ref{sec:#1}}
\newcommand\fig[1]{Fig.~\ref{fig:#1}}

\newcommand\tab[1]{Tab.~\ref{tab:#1}}

\algnewcommand\algorithmicforeach{\textbf{for each}}
\algdef{S}[FOR]{ForEach}[1]{\algorithmicforeach\ #1\ \algorithmicdo}

\definecolor{iccvblue}{rgb}{0.21,0.49,0.74}
\usepackage[pagebackref,breaklinks,colorlinks,allcolors=iccvblue]{hyperref}

\def\paperID{1760} %
\def\confName{ICCV}
\def\confYear{2025}

\usepackage{amsmath}

\long\def\comment#1{}

\makeatletter
\DeclareRobustCommand{\pdot}{\mathbin{\mathpalette\pdot@\relax}}
\newcommand{\pdot@}[2]{%
  \ooalign{%
    $\m@th#1\circ$\cr
    \hidewidth$\m@th#1\cdot$\hidewidth\cr
  }%
}
\makeatother

\title{COSTARR: Consolidated Open Set Technique with Attenuation for Robust Recognition}

\author{Ryan Rabinowitz$^{1}$\thanks{Equal contribution.} \quad
Steve Cruz$^{2}$\footnotemark[1] \quad
Walter Scheirer$^{2}$ \quad
Terrance E. Boult$^{1}$ \quad
\vspace{0.5em}\\
\normalsize $^1$University of Colorado Colorado Springs, USA \\
\normalsize $^2$University of Notre Dame, USA \\
{\tt\small \{rrabinow,tboult\}@uccs.edu \{stevecruz,walter.scheirer\}@nd.edu} \\
}

\begin{document}

\maketitle

\begin{abstract}
Handling novelty remains a key challenge in visual recognition systems.
Existing open-set recognition (OSR) methods rely on the familiarity hypothesis, detecting novelty by the absence of familiar features.
We propose a novel attenuation hypothesis: small weights learned during training attenuate features and serve a dual role—differentiating known classes while discarding information useful for distinguishing known from unknown classes.
To leverage this overlooked information, we present COSTARR, a novel approach that combines both the requirement of familiar features and the lack of unfamiliar ones.
We provide a probabilistic interpretation of the COSTARR score, linking it to the likelihood of correct classification and belonging in a known class.
To determine the individual contributions of the pre- and post-attenuated features to COSTARR's performance, we conduct ablation studies that show both pre-attenuated deep features and the underutilized post-attenuated Hadamard product features are essential for improving OSR.
Also, we evaluate COSTARR in a large-scale setting using ImageNet2012-1K as known data and NINCO, iNaturalist, OpenImage-O, and other datasets as unknowns, across multiple modern pre-trained architectures (ViTs, ConvNeXts, and ResNet).
The experiments demonstrate that COSTARR generalizes effectively across various architectures and significantly outperforms prior state-of-the-art methods by incorporating previously discarded attenuation information, advancing open-set recognition capabilities.
\end{abstract}

\section{Introduction}
\label{sec:introduction}

\begin{figure}[t!]
  \centering
  \subfloat[COSTARR Processing\label{fig:teaser:overview}]{\includegraphics[width=\columnwidth]{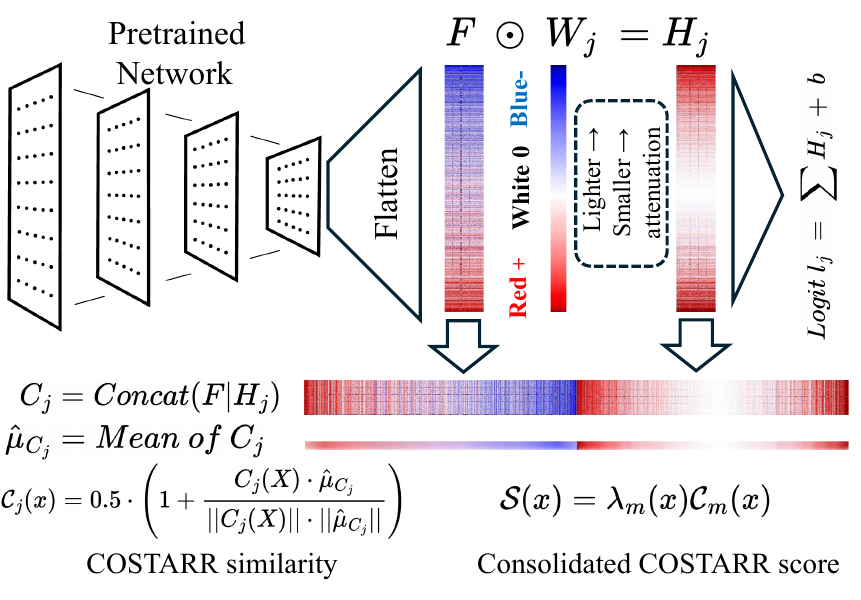}} \\
  \subfloat[OSCR Hiera-H w/NINCO\label{fig:teaser:oscr-hiera}]{\includegraphics[width=.4\linewidth]{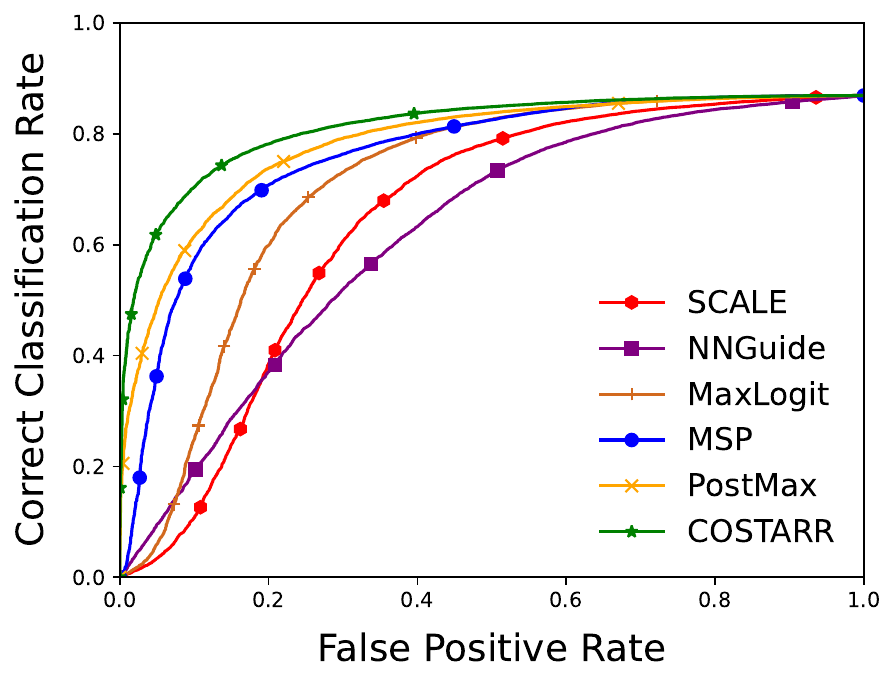}}
  \subfloat[OSCR ViT-H w/NINCO\label{fig:teaser:oscr-vit}]{\includegraphics[width=.4\linewidth]{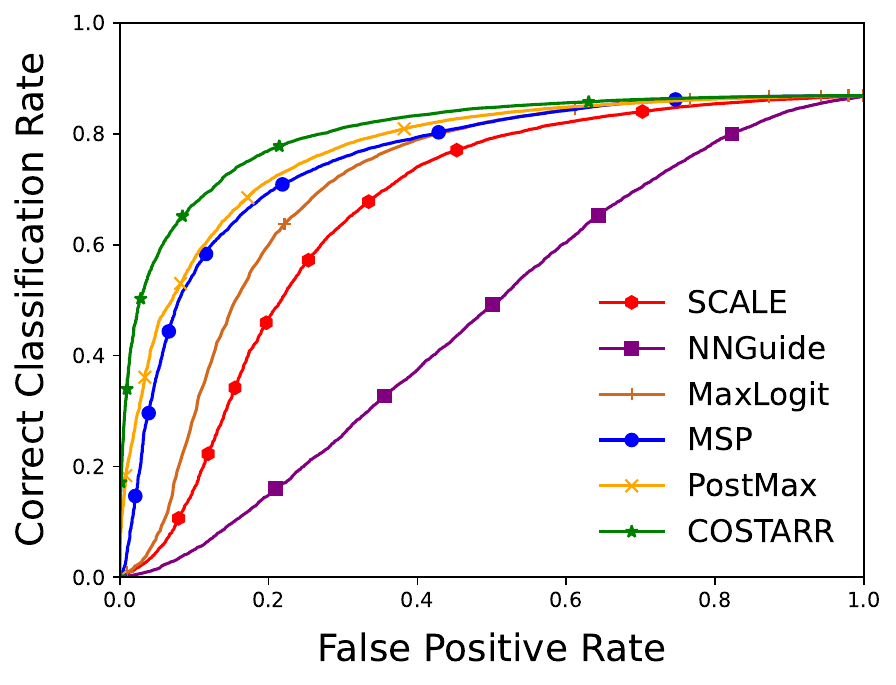}}
  \Caption[fig:teaser]{Overview}{
  \subref{fig:teaser:overview} The Hadamard product $H$ of the feature vector $F$ and the weight vector $W$ for class $j$ is combined with the per-class mean comparison ${\hat \mu}_{C_j}$ to compute the COSTARR similarity score ${\cal C}_j(x)$. This similarity is then scaled by the normalized logit $\lambda_m$ to produce the final COSTARR score ${\cal S}(x)$. See \sec{approach} for details. Our attenuation hypothesis motivates the need for both components. \subref{fig:teaser:oscr-hiera} and \subref{fig:teaser:oscr-vit} demonstrate that COSTARR outperforms current state-of-the-art algorithms, including PostMax, in terms of Open Set Classification Rate (OSCR), highlighting how our novel approach effectively leverages consolidated information to advance the state of the art.}
\end{figure}

\begin{figure*}[t!]
  \centering
  \includegraphics[width=1\textwidth]{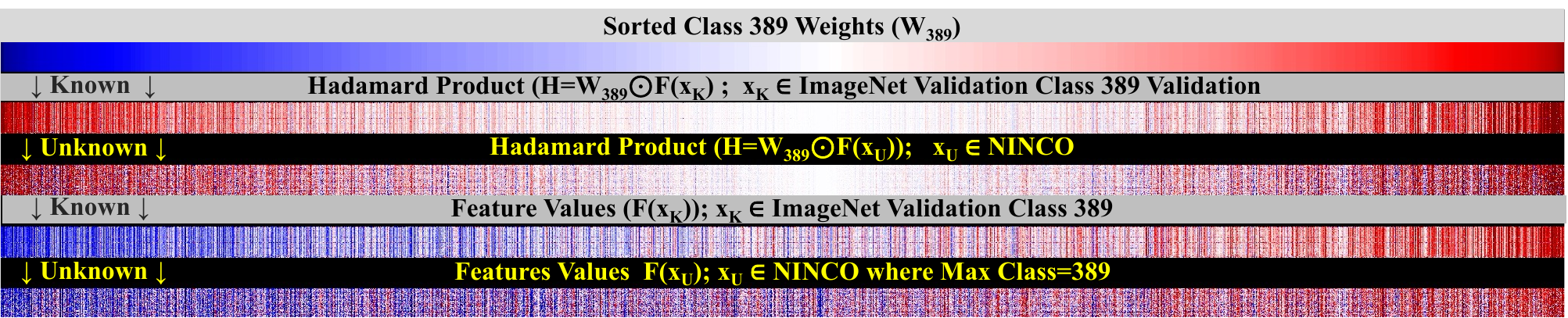}
  \Caption[fig:colorbars]{Intuition behind the Attenuation Hypothesis}{
  The logit for class $j$ is computed as the dot product between its weight vector $W_j$ and the input features $F$, but features associated with low weights are attenuated and thus ignored by this class's projection. The top color bar shows the weights ($W$) for ImageNet2012 class 389 from Hiera-H's classification layer, sorted from low to high (left to right); the white region indicates where weights approach zero. This sorting index is used to order all subsequent color bars. The second bar displays the Hadamard product ($H$) for a known input $x_k$, and the third shows the same for an unknown input $x_u$, for which class 389 produced the max logit. Each column represents a single feature dimension, for 40 different images. The fourth and fifth bars represent the feature vectors from the same 40 images and the most confident unknown samples from NINCO \cite{bitterwolf2023or}, respectively. The consistent low-saturation regions in the weight and Hadamard bars demonstrate that many feature dimensions are attenuated before contributing to classification, regardless of which features are present. Notably, while the feature vectors of NINCO samples are visually dissimilar to those from the validation set, attenuation in the Hadamard and logit computations diminishes their effect. The average logit for the known samples is 8.959, while for the unknowns it is 8.143-making them difficult to distinguish by logit alone. In contrast, the average COSTARR score for samples from known class 389 is 0.647, compared to 0.369 for the unknowns, providing a clearer separation. (Figure best viewed in color.)
}
\end{figure*}

Dealing with unknown inputs in a recognition system is an important and widely recognized problem, which can be formalized as Open-Set Recognition (OSR) \cite{scheirer2012toward}.
The recently introduced familiarity hypothesis \cite{dietterich2022familiarity} frames existing DNN open-set systems as ``detecting the absence of familiar [deep] learned features rather than the presence of novelty''.
Our novel COSTARR approach (Fig.~\ref{fig:teaser}) stems from us asking {\em why do systems implicitly focus on only familiar features} and presents our novel \textit{attenuation hypothesis} to explain it.  

Let $F(x)$ represent the pre-attenuation deep features extracted from input $x$.
Let $W_j$ be the weight vector for class $j$. 
The Hadamard product $H_j = F(x) \odot W_j$ represents the pointwise multiplication that occurs in the classification layer and is a post-attenuated feature.
These two sides of the attenuation, pre- and post- attenuation vectors,  form the foundation of our hypothesis detailed in \sec{approach}.

In OSR, there exists an inherent trade-off between accurately classifying knowns and effectively rejecting unknowns, with each objective benefiting from either pre-- or post--attenuation representations.
We empirically demonstrate that by not ignoring either features, i.e., by incorporating both pre--attenuation and post--attenuation features into our overall confidence measure, we improve a DNN's ability to recognize known classes among unknown samples.
Ablations show that we need to consolidate information from both sides of the attenuation to provide robust recognition.

We formally introduce the attenuation hypothesis in \sec{atten}, but we use \fig{colorbars} here to build intuition and help readers visualize the underlying concept.
The figure illustrates how weights and features contribute to the computation of the maximum classification logit.
In many dimensions, the corresponding weights attenuate the associated features, resulting in a marginal contribution to the final logit.
This attenuation can allow an unknown input to have high confidence for a known class.
Due to the design of fully connected layers, such dimensions are uninformative for the current class but likely play important roles in distinguishing other classes.
However, unknown inputs often exhibit unexpected activation patterns in these attenuated dimensions.
By incorporating both $F$ and $H$, our novel COSTARR score effectively captures these discrepancies.
COSTARR is applicable to any pre-trained network containing a classification layer and incurs minimal computational and storage overhead.

The contributions of this paper are as follows:
\begin{itemize}
\item A novel \textit{attenuation hypothesis} explaining the \textbf{necessary} roles of pre- and post-attenuated features in OSR systems.
\item COSTARR: an efficient, state-of-the-art OSR post-processing algorithm that combines pre-attenuation features, Hadamard product-based post-attenuation features, and logits for robust recognition. Code publicly available.\footnote{\url{https://github.com/Vastlab/COSTARR}}
\item Ablation studies confirm the necessity of both similarity components proposed in the attenuation hypothesis.
\item Sec~\ref{sec:formal} formalizes COSTARR and provides an interpretation of it as a probability estimate that an input is from a known class and the selected class is correct.
\item Experiments on ImageNet-1K across five leading architectures and multiple unknown datasets show statistically significant improvements on several OSR metrics.
\end{itemize}

\section{Related Work}
\label{sec:related}

Numerous works \cite{dhamija2018reducing, Zhou_2021_CVPR,Kong_2021_ICCV,Yoshihashi_2019_CVPR,Neal_2018_ECCV, huang2022class, moon2022difficulty, zhang2022learning, lu2022pmal, chen2021adversarial, guo2021conditional, bao2021evidential,sun2020conditional, miller2021class, chen2020learning} have trained Deep Neural Networks (DNNs) to mitigate the problems posed by unknowns, however, they are categorically different from ours.
We focus exclusively on improving open-set recognition (OSR) after the training process.
Accordingly, we consider other work that focuses on improving OSR for pre-trained networks.

\subsection{Open-Set Post-processors}

OpenMax \cite{bendale2016openmax}, the foundational work for adapting DNNs to open-set, sought to overcome the overconfidence of DNNs trained with softmax activations. 
Using mean activation vectors from the penultimate layer, OpenMax fit Weibull distributions for each class using Extreme Value Theory (EVT).
At test time, OpenMax converted sample distance from the top predicted classes to probabilities and, using a softmax-like operation, revised these probabilities into a vector summing to 1 with an explicit probability of unknown.
Another open-set DNN adaptation, the Extreme Value Machine (EVM) \cite{rudd2017evm}, fits Weibull models between crucial in-class samples (extreme vectors) to negative samples from other classes.
Using the distance of a sample to extreme vectors, the EVM generated probabilities of which class the sample belongs to.
By selecting extreme vectors, the work aimed to minimize open space risk. Both were interesting but experimentally they failed to produce algorithms much better than just using a good close-set classifier \cite{vaze2022openset}.

\comment{
We conjecture these techniques have two distinct pitfalls: revising the top known classifications based on distance and fitting the Weibull models.
DNNs for classification are often trained with softmax, resulting in class-feature distributions that are elongated \cite{dhamija2018reducing}, so it is not intuitive as to why distance from a class mean will be a better predictor for closed set classification.
Secondly, fitting Weibull models (be it from extreme vectors or from mean activation vectors) makes classes that may have slightly different feature distributions compete.
By EVT Type 1 \cite{Gumbel1954StatisticalTO}, a Weibull fit with a ``tighter" feature distribution will have a steeper probability dropoff than other classes with ``looser" feature distributions.
This can make two equal ``distances" in feature space appear dissimilar due to them being normalized by different EVT models.
}

Multiple works have proposed using either Maximum Logit  (MaxLogit) \cite{vaze2022openset,hendrycks2022scaling} or Maximum Softmax Probability (MSP) \cite{hendrycks17baseline,vaze2022openset} as a baseline for OSR. 
These methods are intuitive and widely applicable, as they simply threshold on existing network outputs.
However, PostMax \cite{cruz2025operational}, the recent state-of-the-art for large-scale OSR, demonstrated that there is still much room for improvement.

The current state of the art is PostMax \cite{cruz2025operational} which uses EVT and feature magnitudes to normalize logit ``confidence".
Its predictions are transformed into a probability space, similar to SoftMax, except that there is only one probability per sample rather than a probability for each class.
While their normalizing with feature magnitudes was effective, it relies on the observation that unknowns have a higher feature magnitude than knowns, which contradicts observations of others \cite{dhamija2018reducing, vaze2022openset} and need not be true.
In contrast, we make no such reliance and compare the similarity between known class centers and test samples to capture the information that is generally ignored.
We note that PostMax did not offer a compelling reason for its improvement, just an observation that feature magnitudes were different between known and unknowns.    
Based on our analysis  we  hypothesize that PostMax's improvement is because that, consistent with the \textit{attenuation hypothesis}, they exploit raw feature magnitudes thus their normalization uses some of the ignored information compared, albeit in a weak way.  

\subsection{Out-of-Distribution Post-processors}

Distinct from OSR, Out-of-Distribution (OOD) only focuses on the detection of samples as in-distribution or OOD.
Various approaches have been proposed \cite{liu2020energy, sun2021react, wang2022vim, liu2023gen, rajasekaran2024combood, park2023nearest, liu2024fast, xu2024scaling, fan2024test, tang2024cores} and adopted similar large-scale evaluations, departing from small-scale experimentation in early OOD works \cite{liang2017enhancing, hsu2020generalized} which relied on expensive strategies like reprocessing the inputs.
Given the abundance of methods, OpenOOD \cite{yang2022openood,zhang2023openood} curated a large-scale benchmark to provide an accurate, standardized, and unified evaluation of OOD detection.
We compare our method to others using the large-scale ImageNet-1K benchmark.
Specifically, following prior work \cite{cruz2025operational}, we select the best-performing post-processors: SCALE \cite{xu2024scaling} and NNGuide \cite{park2023nearest}.
SCALE rescales intermediate features, which affects the final output confidence of a network.
This is different from ReAct \cite{sun2021react}, which pruned features outside of a given threshold.
NNGuide \cite{park2023nearest}, drawing off the K-Nearest Neighbor (KNN) ideology, retains a ``bank set" of features from the training set.
Using the bank set and test-time samples, they form a ``guided" score, which reduces over-confidence on far-OOD samples.
Also, we include results provided by the very recent COMBOOD \cite{rajasekaran2024combood}, which combines nearest-neighbor and Mahalanobis distances to form an OOD score.
However, due to difficulties integrating this recent algorithm with our evaluations, we only compare with results from their publication, which did not utilize open-set metrics.
Accordingly, we present this comparison in the supplemental.

\subsection{Metrics}

In addition to introducing PostMax, Cruz \etal{cruz2025operational} proposed the Operational Open-Set Accuracy (OOSA), as a new metric for evaluating open-set recognition systems.
The work was motivated by giving an example of an engineer preparing to deploy an OSR system and the need to determine the overall accuracy of methods before deployment.
OOSA highlights how proper training / validation / testing splits can be used to model real-world settings.
We chose to utilize OOSA as our main evaluation measure as it allows for the prediction of an operational threshold but also recognize other common metrics in the literature and include them for completeness: the Open-Set Classification Rate \cite{dhamija2018reducing} curve  (in our secondary evaluation) and Area Under the Receiver Operating Characteristic curve (in the supplemental material).

\section{Approach}
\label{sec:approach}
Improving algorithm performance is important, but addressing causality and explaining \textbf{why} is even more important for science.
Our approach had two main hypotheses: 1) the novel \textit{attenuation hypothesis} which leads us to combine per-class models using the class mean of both pre- and post-attenuation features, and 2) to maximize usage of training information and provide a consistent probabilistic interpretation, we scaled it using normalized logits to provide the full consolidated model.
We discuss each of these elements.

\subsection{Attenuation Hypothesis} \label{sec:atten}

For class $j$ on input $x$, let Hadamard product $H_j = F(x)\odot W_j$ be the point-wise product of features $F(x)$ with weight vector $W_j$.
The hypothesis states weights that attenuate select features in $W_j$ are important for two competing reasons:

\begin{enumerate} \item To maintain known class accuracy, features used in one class often need to be ignored by others. For class $j$, training yields many small weights that reduce the impact on $H_j$ of some features used in classes $k \ne j$. Similarity to the class $H_j$ mean can measure known class similarity.

\item For unknowns, large feature magnitudes can overwhelm small weight attenuation, causing large logit or softmax values leading to misclassification of unknowns as known. Unknowns can be detected using measures that consider feature magnitude without attenuating weights, e.g., using feature vector $F(x)$ similarity to class $j$ mean so that the added information is not attenuated.
\end{enumerate}

\textbf{The Attenuation Hypothesis:} \textit{
A robust open-set recognition system should integrate both the pre-attenuation features ($F$) and the post-attenuation features ($H$) to optimize the dual objectives of maximizing known class accuracy while minimizing false acceptance of unknown inputs.}

We argue that this hypothesis applies to any network that uses a linear classifier, regardless of the linearity of the input or loss.
We explicitly evaluate the hypothesis in \sec{experiments} across five major networks, including convolutional-based and transformer-based models.
In addition, the hypothesis also helps explain prior work.
The ``familiarity hypothesis,'' \cite{dietterich2022familiarity} suggests that traditional algorithms relying on logits or softmax-based analysis tend to reject unknowns only when they lack familiar features, but fail when encountering inputs with unexpected ones.
We contend this is precisely because such methods rely solely on information in $H$.
Other systems such as OpenMax \cite{bendale2016openmax}, Extreme Value Machine (EVM) \cite{rudd2017evm}, and many prototype-based few-shot open-set learning methods, use only the deep features $F$, and therefore tend to suffer in accuracy on known classes.
To our knowledge only the most recent approach--PostMax \cite{cruz2025operational}, which was the state of the art for large-scale OSR-- attempts to combine information from both $F$ and $H$, albeit in a weak manner and without a clear explanation.

The Attenuation Hypothesis is a key contribution of this paper.
While there may be better ways to leverage the information than our COSTARR approach, we are the first to explain why robust OSR solutions must consider both pre-- and post--attenuation features.

Although not explicitly part of the hypothesis, we also note that the deep network training learns the bias $b$ as part of the final classifier.
This term likely contains useful information that should be incorporated--e.g., through the use of softmax or logits.
Below, we show how to use logits and provide ablations to analyze their importance.

To see this in action, consider the operation performed by the linear layer used to produce classification outputs in a DNN.
As illustrated in \fig{colorbars}, when computing the logit for a single class, the layer's weights attenuate corresponding features--weights approaching zero effectively suppress those features when the Hadamard product is applied.
The resulting values are then summed to produce the logit, which is subsequently scaled to generate the softmax score.
This works well for known inputs. 

The supplement includes a frequency chart showing that many feature dimensions are commonly attenuated to have only a marginal effect.
Analyzing class weights shows that for every feature, there exists at least one class where it is weighted highly and others where it is largely ignored.

When the Hadamard product values are summed, the attenuated features contribute little to the resulting logit or softmax score.
Yet, the response of these features still encodes information about the sample, even if it is not directly useful for the closed-set prediction of the given class--it may be informative for other classes.
Our hypothesis suggests that this discarded information is in fact valuable for distinguishing known from unknown samples, as unknowns tend to exhibit feature responses and Hadamard values that are less consistent with class mean.

\subsection{Per-class Models}
Many models have used per-class information, but none have incorporated both forms of per-class information that we consider. 
While OpenMax and EVM utilize per-class information--such as Mean Activation Vectors and Extreme Vectors--the approaches are weakened by ignoring the additional per-class information in the classification weights.

Recall that PostMax \cite{cruz2025operational} operates by computing the Euclidean norm of feature vectors and dividing the maximum logit by this norm. 
PostMax does not incorporate class-centric information into its normalization factor; regardless of the predicted class, the same Euclidean norm is applied.
While PostMax argues that the method works because unknowns tend to have larger magnitudes, we contend that this observation is just an accidental side effect of the attenuation hypothesis--and for some networks, this assumption does not hold. 
For example, several papers have observed unknowns with smaller magnitudes \cite{dhamija2018reducing, vaze2022openset}, in which case PostMax's normalization would worsen performance.
Moreover, examining figures such as Fig.~7f in \cite{vaze2022openset} reveals that some classes exhibit very different feature magnitude ranges--some smaller than those of unknowns, others larger. 
There appears to be no consistent pattern of unknowns having either larger or smaller magnitudes, suggesting the need for a per-class model instead.

\subsection{Formalizing COSTARR}
\label{sec:formal}
Given the attenuation hypothesis, we aim to define a similarity measure based on the combination of the attenuation-related quantities, $H$ and $F$.
Let $N$ be some pretrained network that produced deep features $F(x)$ on input $x$ and $l_j(x)$ be the logit for class $j$.  Let $K$ be the set of known classes and ${\Omega(x)}$ be the oracle operator that returns the correct class or $-1$ for an unknown class index, and $\Lambda(x)$ be the class with maximum logit from the classifier.  Let $T$ be training data for model building, with $T_j$ the subset correctly classified as class $j$ -- ideally we would use separate data from the network training but for the pretrained networks herein we use the ImageNet training data and use the val set for knowns for testing.
We define a globalized normalized logit (GNL) function to normalize our logits using a global min-max normalization based on training to ensure all normalized logits are in the [0,1] range:
\begin{equation}
    GNL(l)=\max(0,\min(1,\frac{l-l_{tmin}}{l_{tmax}-l_{tmin}}))
\end{equation}  
where $l_{tmin},l_{tmax}$ are the minimum and maximum logits overall classes for correctly classified inputs  $x\in T, \Lambda(x)=\Omega(x)$. Note we avoid per-instance normalization or the use of softmax since there is important information in the relative scale of logits, e.g. smaller versions of object will often have smaller logits. 
At inference time we let
\begin{equation}
 \lambda_m(x)=\max_j \; GNL(l_j(x))
 \end{equation}
be the maximum overall all classes GNL of the logit on input $x$, and note since this is a global linear transform, it does not change which class is maximum, so $\Lambda(x)=m$ is the associated class index.  

We formalize COSTARR scoring through the following steps.  First, our concatenated attenuation-related vector is:
\begin{equation}
  C_{j}(x) =  Concat(F(x),H_j)
\end{equation}
\noindent where $H_j= F(x) \odot W_{j}$ is the Hadamard product, i.e., a pointwise product of weight vector $W_j$ and feature vector $F(x)$ for class $j$ on input $x$.  
Let  ${\hat \mu}_{C_j}$ be the pre-computed mean of ${C_j}(x) \forall x\in T_j$.   Given the co-staring role of both $F$ and $H$  in this we define our COSTARR similarity ${\cal C}_j(x)$ for class $j$ and consolidated COSTARR score ${\cal S}(x)$ as:
\begin{align}
    {\cal C}_j(x) &= 0.5 \cdot \left({1 + \frac{{C_j}(X) \cdot {\hat \mu}_{C_j}}{||{C_j}(X)||  \cdot ||{\hat \mu}_{C_j}||} }\right)\label{eq:costarsim}  \\
    {\cal S}(x) &= \lambda_m(x)  {\cal C}_m(x) 
\end{align}
where the $0.5$ in Eq.~\ref{eq:costarsim} rescales the similarity into [0,1]. 
In processing, we select the class $m=\Lambda(x)$ with maximum GNL logit $\lambda_m$ and then use the precomputed mean for the consolidated attenuation-related vector of class $m$ to compute the similarity.  Thus we interpret the COSTARR similarity as an approximation of $P(x\in K | \Lambda(x)=m)$, i.e. the conditional probability the input is from a known class given we observed the input is from class $m$.  If we assume that our GNL normalized max logit estimates the probability for that class, which is a good approximation of the correct class for known inputs, i.e. $ \lambda_m(x)\approx P(\Lambda(x)=m) $, then we can interpret the COSTARR score as 
\begin{align}
{\cal S}(x) &\approx  P(\Lambda(x)=m) \cdot P((x \in K )| (\Lambda(x)=m))  \label{eq:sexpand}\\
&=  P(\Lambda(x)=m)  \cdot \frac{P ((x \in K)\cap (\Lambda(x)=m))}{P(\Lambda(x)=m)} \label{eq:multlogit} \\
&= {P ((x \in K)\cap (\Lambda(x)=m))} \\
&\approx {P ((x \in K)\cap (\Omega(x)=m))} \label{eq:finaleq}
\end{align}
\noindent where Eq.~~\ref{eq:multlogit} follows from \ref{eq:sexpand} from the definition of conditional probability and \ref{eq:finaleq} uses the assumption that on known inputs our classifier is sufficiently accurate.  Thus the novel COSTARR score is an approximation of the probability that an input is both known and of the correct class. Hence, it can be thresholded to separate correctly classified knowns from unknown or misclassified ones. 

Since we condition our algorithm--and the associated probabilities--on the max logit class, a limitation of this approach is that the probability approximation can fail on inputs from known classes when the pretrained network misclassifies them; that is, when $ x \in K \hbox{ and }  \Lambda(x) \ne \Omega(x)$. 
In such cases, the mean of the incorrect class is used, resulting in a lower COSTARR score.
Consequently, the model is more likely to declare the input as unknown.
From an operational perspective, this ``limitation''--rejecting incorrectly classified known inputs--may actually be beneficial, as such misclassifications are likely to cause downstream errors. 

This analysis also explains why we multiply by the normalized logit: in Eq.~\ref{eq:multlogit}, it cancels out the denominator that arises in the expansion of the conditional probability.
Without this term, variations in per-class accuracy (i.e., errors in ${P(\Lambda(x)=m))}$) or poor calibration would directly affect the similarity score.
By canceling this out, we ensure that the sorted order of scores remains reliable for thresholding, even if the scores themselves are not well calibrated. 

While we define a particular similarity above, the attenuation hypothesis does not specify exactly how to compute similarity between the related terms.
Although a global monotonic transformation of $\lambda(x)$ does not affect sorted order, per-instance normalization or changes to the similarity function can change it, which impacts open-set performance.
In our ablations, we demonstrate the importance of considering both the attenuated $H$  and the original features $F$, validating the attenuation hypothesis.    
We do not claim that \ref{eq:finaleq} is the optimal combination.
We expect ``optimal'' will depend on the network and the data used, as both impact $F$ and $H$ and the accuracy of logits on which we condition. 
Practitioners applying this approach may want to explore alternate combinations, normalizations, or optimizations, particularly after fine-tuning models for their specific tasks.
Nonetheless, a key contribution of this paper is the attenuation hypothesis itself.
Those aiming to optimize performance should ensure both $F$ and $H$ are incorporated into their methods.

\section{Experiments}
\label{sec:experiments}

The theory behind COSTARR was presented above, but experimental validation is critical to show the underlying assumptions hold in the real world. To effectively evaluate our OSR performance in a large-scale setting, we adapt the protocol recently established by Cruz \etal{cruz2025operational}, with their provided code. 
Their protocol consists of large-scale datasets and modern architectures and allows the prediction of an operational threshold closely resembling a real-world system.
Additionally, we conduct ablations to understand the performance impact of our approach and its components.

\subsection{Details}

\subsubsection{Datasets}
We utilize the large-scale ImageNet2012 \cite{ILSVRC15} as our known class training and test sets.
For OOSA threshold prediction, we employ ImageNetV2 \cite{recht2019imagenet} as our known validation set.

Bitterwolf \etal{bitterwolf2023or} recently showed that the 21K-P Open-Set splits \cite{vaze2022openset} commonly used for OOD/unknowns evaluation contain over 40\% overlap with ImageNet-1K, raising concerns about in-distribution contamination.
Using such splits to estimate operational thresholds or evaluate performance can be misleading—we elaborate on this in the supplemental with supporting examples.
To avoid this issue, we follow the protocol by Cruz \etal{cruz2025operational}, but replace their unknowns surrogate with 10K images from OpenImage-O, which has no significant overlap.
For completeness, we also report results on 21K-P splits in the supplement to match PostMax’s evaluation.

At test time, we use 10K images from the plant classes in iNaturalist (iNat) \cite{Horn_2018_CVPR} and the remaining 7.6K images from OpenImage-O (Open-O) \cite{wang2022vim} as unknowns.
We also include the NINCO \cite{bitterwolf2023or} (5.8K images) and Textures (Text) \cite{Cimpoi_2014_CVPR} (5.1K images) datasets as unknowns.
Bitterwolf \etal{bitterwolf2023or} reported significant contamination in several commonly used datasets, including Places \cite{zhou2017places} (59.5\%) and Textures \cite{Cimpoi_2014_CVPR} (20.0\%).
Using contaminated datasets for OSR is problematic, as high confidence on mislabeled unknowns can compromise evaluation; properly labeled unknowns typically yield better performance.
In the supplemental, we provide examples where contaminated samples closely resemble ImageNet-1K training classes.
While Textures and OpenImage-O each have ~20\% overlap, we include them to increase dataset diversity, with further analysis provided in the supplemental.
However, we caution against over-interpreting results on these contaminated datasets.

\subsubsection{Architectures}

By utilizing ImageNet2012 \cite{ILSVRC15}, we can select well-studied pre-trained architectures for our evaluation.
We evaluate performance on the traditional ResNet-50 \cite{he2016deep} and two recent state-of-the-art architectures pre-trained on ImageNet2012 only: ConvNeXtV2-H \cite{woo2023convnext} and Hiera-H \cite{ryali2023hiera}. We also use a Vision Transformer trained with masked-auto-encoder  ViT/MAE\cite{he2022masked} and the original ConvNeXt \cite{liu2022convnet}.
For each network, we extract features from the penultimate layer and the added special code to extract the Hadamard product.

\subsubsection{Metrics}
Our main results utilize the recent Operational Open-Set Accuracy (OOSA) \cite{cruz2025operational}, using code obtained from the authors.
OOSA evaluates algorithm performance using the predicted threshold from validation.
Additionally, we also utilize the traditional OSR metric, the Open-Set Classification Rate (OSCR) \cite{dhamija2018reducing} as area under the curve in tables here and plots in the supplemental. 
In the supplemental, we also include Area Under the Receiver Operator Curve (AUROC), commonly used in the OOD literature but stress that AUROC should be interpreted with caution as it is generally incoherent for comparisons of algorithms \cite{hand_when_2013}.

\begin{table}[t!]
\begin{tabular}{c?c?cccc}
\multicolumn{1}{c?}{\rotatebox[origin=c]{90}{{\textbf{Arch}}}} & \multicolumn{1}{c?}{\textbf{Method}} & \textbf{iNat} & \textbf{NINCO} & \textbf{Open-O} & \textbf{Text} \\ \noalign{\hrule height 1.0pt}
\multirow{6}{*}{\rotatebox[origin=c]{90}{{ResNet-50}}} & SCALE & 0.635 & 0.555 & 0.581 & 0.628 \\
 & NNGuide & 0.633 & 0.445 & 0.571 & 0.518 \\
 & MaxLogit & 0.705 & 0.641 & 0.679 & 0.636 \\
 & MSP & 0.745 & 0.675 & 0.719 & 0.676 \\
 & PostMax & 0.762 & 0.623 & 0.730 & 0.696 \\
 & COSTARR & \textbf{0.804} & \textbf{0.693} & \textbf{0.773} & \textbf{0.749} \\ \hline
\multirow{6}{*}{\rotatebox[origin=c]{90}{{ConvNeXt-L}}} & SCALE & 0.681 & 0.650 & 0.668 & 0.679 \\
 & NNGuide & 0.601 & 0.513 & 0.615 & 0.505 \\
 & MaxLogit & 0.744 & 0.692 & 0.718 & 0.711 \\
 & MSP & 0.778 & 0.707 & 0.752 & 0.722 \\
 & PostMax & 0.820 & 0.714 & 0.798 & 0.761 \\
 & COSTARR & \textbf{0.837} & \textbf{0.744} & \textbf{0.814} & \textbf{0.778} \\ \hline
  \multirow{6}{*}{\rotatebox[origin=c]{90}{{ConvNeXtV2-H}}} & SCALE & 0.777 & 0.713 & 0.752 & 0.767 \\
 & NNGuide & 0.748 & 0.628 & 0.700 & 0.643 \\
 & MaxLogit & 0.800 & 0.732 & 0.775 & 0.773 \\
 & MSP & 0.808 & 0.734 & 0.782 & 0.756 \\
 & PostMax & 0.850 & 0.755 & 0.830 & 0.796 \\
 & COSTARR & \textbf{0.856} & \textbf{0.756} & \textbf{0.835} & \textbf{0.802} \\ \hline
\multirow{6}{*}{\rotatebox[origin=c]{90}{{ViT-H}/MAE-H}} & SCALE & 0.758 & 0.689 & 0.710 & 0.750 \\
 & NNGuide & 0.639 & 0.550 & 0.672 & 0.541 \\
 & MaxLogit & 0.797 & 0.721 & 0.750 & 0.765 \\
 & MSP & 0.812 & 0.737 & 0.777 & 0.761 \\
 & PostMax & 0.865 & 0.749 & 0.841 & 0.808 \\
 & COSTARR & \textbf{0.876} & \textbf{0.780} & \textbf{0.851} & \textbf{0.820} \\ \hline
\multirow{6}{*}{\rotatebox[origin=c]{90}{{Hiera-H}}} & SCALE & 0.732 & 0.682 & 0.684 & 0.742 \\
 & NNGuide & 0.798 & 0.644 & 0.770 & 0.670 \\
 & MaxLogit & 0.796 & 0.723 & 0.747 & 0.768 \\
 & MSP & 0.822 & 0.744 & 0.785 & 0.772 \\
 & PostMax & 0.870 & 0.758 & 0.850 & 0.818 \\
 & COSTARR & \textbf{0.879} & \textbf{0.788} & \textbf{0.857} & \textbf{0.826}
\end{tabular}
\Caption[tab:oosa]{Operational Open-Set Accuracy}{The mean OOSA ($\uparrow$) of all methods. To predict an operational threshold, we validate the methods using ImageNetV2 \cite{recht2019imagenet} ($10K$ images) as knowns and OpenImage-O \cite{wang2022vim} ($10K$ images) as unknowns. Then, each method's threshold is deployed and tested on five ILSVRC2012 \emph{val} \cite{ILSVRC15} splits (each containing $10K$ images) and specified unknowns. OSR is performed on extractions from various pre-trained architectures. COSTARR, our novel algorithm, has the best scores (\textbf{bold}) for each respective architecture and unknowns dataset.}
\end{table}

\subsection{Results}

We compared COSTARR with recent approaches: PostMax \cite{cruz2025operational} - current state-of-the-art for OSR, Maximum Logit (MaxLogit) \cite{hendrycks2022scaling, vaze2022openset}, and Maximum Softmax Probability (MSP) \cite{hendrycks17baseline, vaze2022openset}.
Additionally, we include recent state-of-the-art OOD methods (as determined by the OpenOOD Benchmark leaderboard \cite{yang2022openood,zhang2023openood}): SCALE \cite{xu2024scaling} and NNGuide \cite{park2023nearest}. 
Since COMBOOD \cite{rajasekaran2024combood} is not yet integrated within OpenOOD, we cannot reproduce its results; however, we include results from their paper in the supplemental. 

As shown in \tab{oosa}, COSTARR outperforms other methods on OOSA, a powerful metric used to measure the deployment characteristics of Open-Set algorithms. Notably, NINCO \cite{bitterwolf2023or}, a purpose-built OOD dataset designed to avoid contamination with ImageNet2012, provides a powerful point of comparison at which COSTARR excels. 
To ensure the results are not just random effects, dataset-dependent, architecture-specific, or validation tuning results, we computed Wilcoxon signed rank test, as implemented in scipy, with Bonferroni correction across Post-Max's five different splits of the validation data with each architecture and algorithm -- see supplemental for more details. 
Note this is different than the t-tests used in \cite{cruz2025operational}, because the data is reused in ways that likely violate the independence needed for t-tests.
All statistical claims in the paper use this test process. 
Since OOSA computes thresholds directly, there are {\em NO free/tuned parameters} in these experiments.   
The result of the statistical testing is that COSTARR is statistically significantly better ($p< 10^{-5} $) on each architecture separately, as well as very significantly ($p< 10^{-6}$) across all architectures combined.  
Additionally, while Places \cite{zhou2017places}, 21K-P Easy/Hard \cite{vaze2022openset}, and SUN \cite{xiao2010sun} have significant ImageNet2012 contamination, COSTARR still performs significantly ($p<10^-3)$) better than all methods except PostMax, but was never statistically worse. 
We include those datasets in the supplemental, with a discussion about data contamination and statistical testing. 

\begin{table}[t!]
\begin{tabular}{c?c?cccc}
\multicolumn{1}{c?}{\rotatebox[origin=c]{90}{{\textbf{Arch}}}} & \multicolumn{1}{c?}{\textbf{Method}} & \textbf{iNat} & \textbf{NINCO} & \textbf{Open-O} & \textbf{Text} \\ \noalign{\hrule height 1.0pt}
\multirow{6}{*}{\rotatebox[origin=c]{90}{{ResNet-50}}} & SCALE & 0.621 & 0.499 & 0.545 & 0.598 \\
 & NNGuide & 0.596 & 0.419 & 0.551 & 0.557 \\
 & MaxLogit & 0.682 & 0.625 & 0.661 & 0.609 \\
 & MSP & 0.720 & 0.664 & 0.703 & 0.664 \\
 & PostMax & 0.743 & 0.624 & 0.730 & 0.720 \\
 & COSTARR & \textbf{0.773} & \textbf{0.699} & \textbf{0.760} & \textbf{0.755} \\ \hline
\multirow{6}{*}{\rotatebox[origin=c]{90}{{ConvNeXt-L}}} & SCALE & 0.669 & 0.628 & 0.636 & 0.657 \\
 & NNGuide & 0.594 & 0.462 & 0.616 & 0.453 \\
 & MaxLogit & 0.729 & 0.677 & 0.695 & 0.693 \\
 & MSP & 0.771 & 0.717 & 0.751 & 0.731 \\
 & PostMax & 0.800 & 0.725 & 0.792 & 0.770 \\
 & COSTARR & \textbf{0.815} & \textbf{0.761} & \textbf{0.809} & \textbf{0.793} \\ \hline
 \multirow{6}{*}{\rotatebox[origin=c]{90}{{ConvNeXtV2-H}}} & SCALE & 0.773 & 0.712 & 0.748 & 0.771 \\
 & NNGuide & 0.759 & 0.639 & 0.719 & 0.653 \\
 & MaxLogit & 0.794 & 0.735 & 0.771 & 0.779 \\
 & MSP & 0.810 & 0.754 & 0.791 & 0.782 \\
 & PostMax & 0.834 & 0.773 & 0.827 & 0.805 \\
 & COSTARR & \textbf{0.838} & \textbf{0.782} & \textbf{0.831} & \textbf{0.814} \\ \hline
\multirow{6}{*}{\rotatebox[origin=c]{90}{{ViT-H}/MAE-H}} & SCALE & 0.749 & 0.661 & 0.679 & 0.744 \\
 & NNGuide & 0.646 & 0.469 & 0.670 & 0.442 \\
 & MaxLogit & 0.790 & 0.709 & 0.730 & 0.764 \\
 & MSP & 0.816 & 0.755 & 0.785 & 0.782 \\
 & PostMax & 0.846 & 0.772 & 0.839 & 0.822 \\
 & COSTARR & \textbf{0.854} & \textbf{0.803} & \textbf{0.848} & \textbf{0.835} \\ \hline
\multirow{6}{*}{\rotatebox[origin=c]{90}{{Hiera-H}}} & SCALE & 0.718 & 0.643 & 0.633 & 0.730 \\
 & NNGuide & 0.793 & 0.620 & 0.781 & 0.669 \\
 & MaxLogit & 0.783 & 0.706 & 0.715 & 0.765 \\
 & MSP & 0.822 & 0.760 & 0.788 & 0.792 \\
 & PostMax & 0.846 & 0.784 & 0.842 & 0.825 \\
 & COSTARR & \textbf{0.856} & \textbf{0.812} & \textbf{0.850} & \textbf{0.838}
\end{tabular}
\Caption[tab:AUOSCR]{Area Under Open Set Classification Rate Curve}{The AUOSCR ($\uparrow$) of all methods. To compute, we tested methods using ILSVRC2012 \emph{val} \cite{ILSVRC15} ($50K$ images) as knowns and specified unknowns. The best scores for each respective architecture and unknowns dataset are in \textbf{bold}.}
\end{table}

Using AUOSCR as a secondary metric, \tab{AUOSCR} shows COSTARR outperforms all methods; again, the differences are statistically significant. In the supplemental we include AUROC tables where again COSTARR is statistically significantly better overall ($p<10^{-3}$).

\begin{table}[t!]
\begin{tabular}{c?c?cccc}
\multicolumn{1}{c?}{\rotatebox[origin=c]{90}{{\textbf{Arch}}}} & \multicolumn{1}{c?}{\textbf{Method}} & \textbf{iNat} & \textbf{NINCO} & \textbf{Open-O} & \textbf{Text} \\ \noalign{\hrule height 1.0pt}
\multirow{6}{*}{\rotatebox[origin=c]{90}{{ResNet-50}}}
 & PostMax & 0.743 & 0.624 & 0.730 & 0.720 \\
 & Hadamard & 0.426 & 0.365 & 0.393 & 0.384 \\
 & Features & 0.746 & 0.681 & 0.736 & 0.740 \\
 & NoLogit & 0.771 & 0.690 & 0.757 & \textbf{0.756} \\
 & CO-SM & 0.757 & 0.693 & 0.743 & 0.725 \\
 & COSTARR & \textbf{0.773} & \textbf{0.699} & \textbf{0.760} & 0.755 \\ \hline
 \multirow{6}{*}{\rotatebox[origin=c]{90}{{ConvNV2}}}
 & PostMax & 0.834 & 0.773 & 0.827 & 0.805 \\
 & Hadamard & 0.834 & 0.772 & 0.829 & 0.807 \\
 & Features & 0.835 & 0.773 & 0.830 & 0.810 \\
 & NoLogit & 0.837 & 0.775 & 0.833 & 0.813 \\
 & CO-SM & 0.832 & 0.778 & 0.823 & 0.807 \\
 & COSTARR & \textbf{0.838} & \textbf{0.782} & \textbf{0.831} & \textbf{0.814} \\ \hline
\multirow{6}{*}{\rotatebox[origin=c]{90}{{Hiera-H}}}
 & PostMax & 0.846 & 0.784 & 0.842 & 0.825 \\
 & Hadamard & 0.853 & 0.807 & 0.848 & 0.833 \\
 & Features & 0.854 & 0.809 & 0.849 & 0.835 \\
 & NoLogit & 0.855 & 0.809 & \textbf{0.850} & 0.836 \\
 & CO-SM & 0.851 & 0.807 & 0.843 & 0.831 \\
 & COSTARR & \textbf{0.856} & \textbf{0.812} & \textbf{0.850} & \textbf{0.838}
\end{tabular}
\Caption[tab:ablations]{Ablation Study}{
AUOSCR scores from our ablation studies; the best scores are in \textbf{bold}. Hadamard refers to a version of COSTARR that uses only the Hadamard product features ($H$) without concatenation, i.e. post-attenuation for similarity and selection. 
Features uses only pre-attenuation features ($F$) for similarity and selection.  
NoLogit shows performance when using the concatenated $F$ and $H$ similarity, with the class selected based on the maximum similarity.
For comparison, we also include PostMax, the prior state of the art. 
We evaluate across three networks (as used in supplemental \fig{weight-histo}: ResNet-50 \cite{he2016deep}, ConvNeXtV2-H \cite{woo2023convnext}, and Hiera-H \cite{ryali2023hiera}.
In all cases, the ablations perform worse than COSTARR, providing empirical evidence that incorporating features typically ignored (discarded or marginalized by the Hadamard product) improves known/unknown differentiation.
The comparison of softmax (CO-SM) \vs logits (COSTARR) further confirms that normalization using logits yields superior performance. 
}
\end{table}

\subsection{Ablations}
We ran an ablation study to examine the key elements of our attenuation hypothesis and to analyze which components of COSTARR contribute to the observed performance gains.
All ablations use AUOSCR as the evaluation metric, as it is not sensitive to threshold selection.

The hypothesis states that both pre-- and post--attenuation features should be used.
Also, it suggests that the final logits, including the learned bias, likely contain valuable information necessary for the probabilistic interpretation to function properly.

We build an ablation version using only COSTARR Similarity (Eq.~\ref{eq:costarsim}), employing either the Hadamard product $H$ or deep features $F$ alone, and excluding the Logit.
Treating the resulting vector as a type of prototype, we select the class with the maximum similarity.
Although there are some architecture and dataset-specific variations, both of these variants perform statistically significantly worse than COSTARR overall ($p<.05$).

We also include an ablation with CO-SM, which uses softmax instead of logits for scaling -- this does not cancel out in the probabilistic interpretation. 
Another ablation demonstrates that even when combining both pre-- and post--attenuation features ($F$ and $H$), the logits still provide additional value.
To test this, we introduce a variant called NoLogit, which uses the full COSTARR similarity from Eq.~\ref{eq:costarsim} to both select the winning class and reject unknowns.
This variant performs slightly better than using either feature alone, but remains weaker than the overall COSTARR.

We were initially surprised by the much lower performance and greater variability of the ablation version using softmax instead of logit (CO-SM).
For some networks and datasets, it outperformed the NoLogit ablation, while for others, it performed worse.
Although both max logit and max softmax select the same class, the difference likely arises from softmax's per-instance normalization, which affects the score and the probabilistic interpretation in Eq.{\ref{eq:finaleq}}.  
Since CO-SM never outperformed COSTARR with logits, we did not explore it further. 

From \tab{ablations}, COSTARR consistently outperforms the ablation variants.
These provide strong evidence in support of our \textit{attenuation hypothesis} and indicate that each component of COSTARR contributes meaningfully to its performance.
Additional discussion is provided in the supplemental, particularly regarding COSTARR's performance.

\section{Discussion and Conclusion}
\label{sec:conclusion}
Exploring our novel \emph{attenuation hypothesis}, we analyzed deep features and classification layer weights, finding that multiple networks rely on classification weights that attenuate general features and why that attenuation is problematic for OSR. 
From our analysis of networks  (see also \fig{weight-histo} in supplement), we showed that each feature has low weight for some classes and high for others, supporting our hypothesis. 

We exploited our \emph{attenuation hypothesis} in our proposed OSR algorithm, COSTARR, which introduces only a constant to test-time complexity.
Our main results on operational open-set accuracy (\tab{oosa}) demonstrate that COSTARR outperforms prior approaches, including the recent PostMax \cite{cruz2025operational}. 
These results are practically better,  statistically significant, and achieved at almost no added computational cost. 
Additionally, results with a secondary OSR evaluation metric, AUOSCR (\tab{AUOSCR}), show statistically significant performance gains over prior approaches. 
The ablations (\tab{ablations}) validated that COSTARR's performance derives from  three components: the use of both pre-- and post-attenuated features as well as a small increase from using logits. 
Through this analysis, we found direct evidence supporting the \emph{attenuation hypothesis} as well as the benefit of using logits and per-class models for OSR.
While the ablations show the effectiveness of each component varies across networks and datasets, COSTARR achieves unprecedented performance across all networks by consolidating all three sources of information.
We leave the exploration of these insights on non-pretrained networks for future work.

Through this exploration, we have advanced the understanding of the weak performance of different open-set classifiers because they discard information useful to either known or unknown samples. COSTARR takes the first steps toward consolidating pre-attenuation and post-attenuation information to improve robust recognition.

{
    \small
    \bibliographystyle{ieeenat_fullname}
    \bibliography{references}
}

\clearpage
\setcounter{page}{1}
\maketitlesupplementary

\begin{figure*}[t!]
  \centering
  \includegraphics[width=1\textwidth]{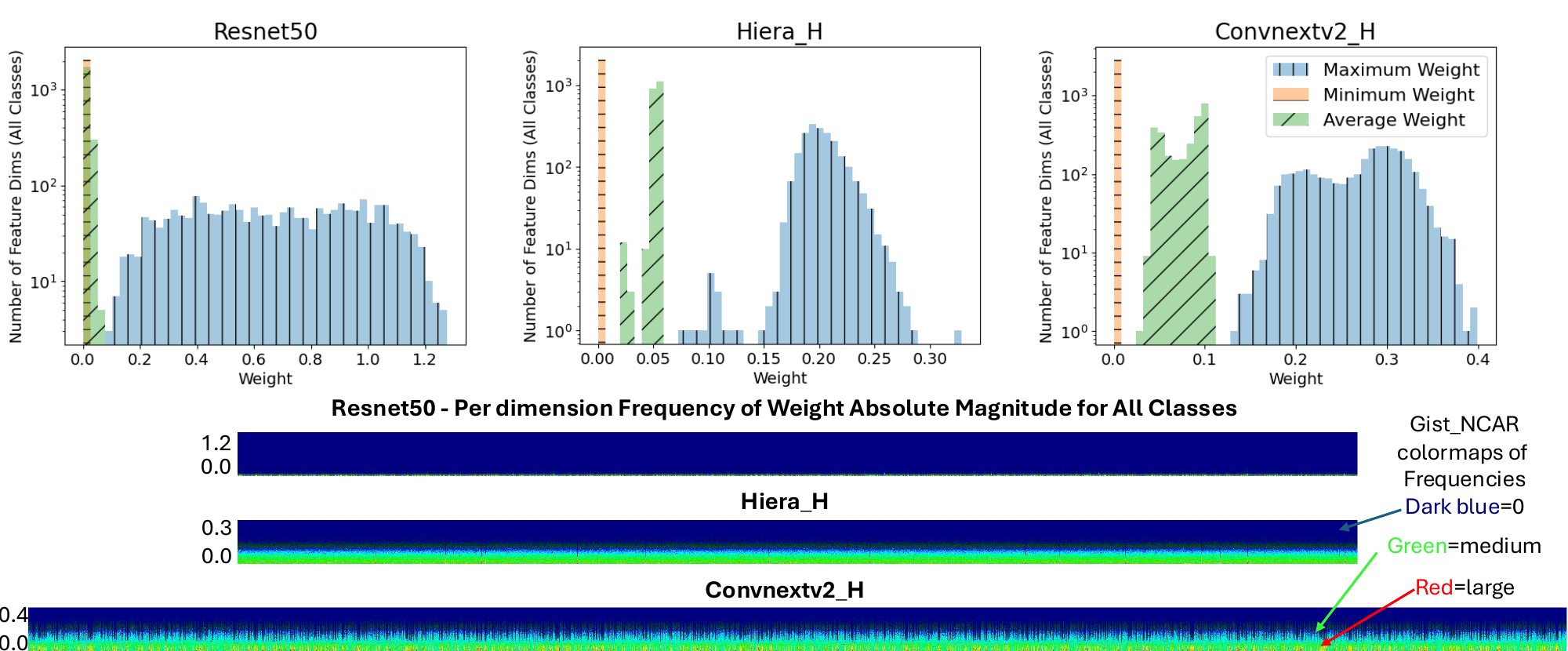}
\Caption[fig:weight-histo]{Histogram of Class-Weights}{The spread of Min (horizontal stripes), Mean (diagonal stripes), and Max (vertical stripes) weights per dimension across all classes illustrate how any given feature can be discarded by the classification layer of a DNN (Hiera-H pictured). 
The Maximum weights are all reasonably above 0, indicating that each feature dimension is useful to some class. 
The minimum weights all approach 0, indicating that there is a class that ignores each feature.
Since the Mean weights tend closer to 0 than to the mean of the max weights, feature dimensions are more often ignored (by having their contribution attenuated) rather than being strong contributors to the final logit.
The bottom three color bars visualize how often this occurs across every class in ImageNet, with the absolute value per-dimension frequency of each weight visualized.
The bars are different widths due to the different feature dimensions from each network.
Each pixel in the 80-pixel-tall columns represents a bin from the range between 0 and the absolute maximum weight of all dimensions and classes; brightness indicates frequency (using the ``gist\_ncar" color scheme).
The concentration near of bright green near the bottom (approaching 0 weight) shows that every feature dimension most often has its contributions to maximum logit confidence attenuated, depending on which class' weights are used.
While these dimensions have been weighted to discriminate between known classes, they ignore information that could help differentiate known from unknown. 
}
\end{figure*}
\section{Extended discussion}

While we stated that probabilist interpretation of COSTARR depends on an accurate classifier, it is interesting to note that in \ref{tab:AUOSCR} the gap over the prior state of the art (PostMax) is largest for ResNet-50, which is the weakest network in terms of closed set accuracy, suggesting there is more to the improvements than just the probabilistic interpretation. 

In the ablation, \ref{tab:ablations},  we see that for ResNet50, the Hadamard only version, with only post-attentive features, did much worse than using only the Features.  We hypothesize that maybe the gains for ResNet50 in other tables are from the concatenation with Features improving class separation beyond that the logit/Hadamard feature can provide.  For Resnet50 with Textures as the unknown, adding the logit did not really improve things, again, maybe suggesting the post-attenuation Hadamard features are weaker in ResNet.

The weakest element of the ablation was the difference between NoLogit and COSTARR, which for two networks there was one dataset where the results without the logit were nearly identical to the overall COSTARR, though the differences were not statistically different. We note that the full COSTARR with logit version is actually cheap at test time since we access only the mean for the max logit class and already have the logit value from that computation.

\section{Metrics Cont.}
For completeness, we include Open-Set Accuracy (OSA) curves in \fig{res-osa}-\ref{fig:hiera-osa}.
Note, the operational performance (OOSA) reported in \tab{oosa}, is indicated by a $\star$ on each curve.
Consistent with the tabular results, COSTARR achieves maximum OSA across all datasets.

In addition to OOSA and AUOSCR metrics included in the main paper, we also evaluated methods using Open-Set Classification Rate (OSCR) curves, shown in \fig{res-oscr}-\ref{fig:hiera-oscr}.
Similar to OOSA (\tab{oosa}) and AUROC in (\tab{auroc}), COSTARR outperforms all methods across all datasets.

\begin{table}[t]
\begin{tabular}{c?c?cccc}
\multicolumn{1}{c?}{\rotatebox[origin=c]{90}{{\textbf{Arch}}}} & \multicolumn{1}{c?}{\textbf{Method}} & \textbf{iNat} & \textbf{NINCO} & \textbf{Open-O} & \textbf{Text} \\ \noalign{\hrule height 1.0pt}
\multirow{6}{*}{\rotatebox[origin=c]{90}{{ResNet-50}}} & SCALE & 0.532 & 0.402 & 0.634 & 0.386  \\
 & NNGuide & 0.513 & 0.385 & 0.647 & 0.357  \\
 & MaxLogit & 0.669 & 0.584 & 0.707 & 0.565  \\
 & MSP & 0.702 & 0.617 & 0.747 & 0.606  \\
 & PostMax & 0.571 & 0.452 & 0.689 & 0.431  \\
 & COSTARR & \textbf{0.718} & \textbf{0.627} & \textbf{0.789} & \textbf{0.627}  \\ \hline
\multirow{6}{*}{\rotatebox[origin=c]{90}{{ConvNeXt-L}}} & SCALE & 0.683 & 0.638 & 0.664 & 0.659  \\
 & NNGuide & 0.583 & 0.447 & 0.662 & 0.450  \\
 & MaxLogit & 0.740 & 0.682 & 0.728 & 0.693  \\
 & MSP & 0.773 & 0.701 & 0.791 & 0.706  \\
 & PostMax & 0.794 & 0.698 & 0.830 & 0.722  \\
 & COSTARR & \textbf{0.801} & \textbf{0.721} & \textbf{0.846} & \textbf{0.730}  \\ \hline
  \multirow{6}{*}{\rotatebox[origin=c]{90}{{ConvNeXtV2-H}}} & SCALE & 0.772 & 0.705 & 0.767 & 0.740  \\
 & NNGuide & 0.713 & 0.597 & 0.745 & 0.601 \\
 & MaxLogit & 0.786 & 0.718 & 0.792 & 0.740  \\
 & MSP & 0.796 & 0.723 & 0.820 & 0.736  \\
 & PostMax & 0.812 & 0.732 & 0.849 & 0.744  \\
 & COSTARR & \textbf{0.819} & \textbf{0.739} & \textbf{0.857} & \textbf{0.752}  \\ \hline
\multirow{6}{*}{\rotatebox[origin=c]{90}{{ViT-H}}} & SCALE & 0.748 & 0.671 & 0.706 & 0.718  \\
 & NNGuide & 0.636 & 0.552 & 0.661 & 0.544  \\
 & MaxLogit & 0.788 & 0.711 & 0.755 & 0.742  \\
 & MSP & 0.794 & 0.717 & 0.809 & 0.727  \\
 & PostMax & 0.826 & 0.732 & 0.861 & 0.761  \\
 & COSTARR & \textbf{0.834} & \textbf{0.759} & \textbf{0.872} & \textbf{0.773}  \\ \hline
\multirow{6}{*}{\rotatebox[origin=c]{90}{{Hiera-H}}} & SCALE & 0.727 & 0.667 & 0.663 & 0.715  \\
 & NNGuide & 0.782 & 0.625 & 0.796 & 0.660  \\
 & MaxLogit & 0.780 & 0.707 & 0.745 & 0.736  \\
 & MSP & 0.815 & 0.739 & 0.814 & 0.756  \\
 & PostMax & 0.825 & 0.740 & 0.864 & 0.764  \\
 & COSTARR & \textbf{0.852} & \textbf{0.782} & \textbf{0.883} & \textbf{0.797} 
\end{tabular}
\Caption[tab:oosa-hard]{Operational Open-Set Accuracy}{The mean OOSA ($\uparrow$) of all methods on the validation from Cruz \etal{cruz2025operational}. To predict an operational threshold, we validate the methods using ImageNetV2 \cite{recht2019imagenet} ($10K$ images) as knowns and 21K-P Hard \cite{vaze2022openset} ($9.8K$ images) as unknowns. Then, each method's threshold is deployed and tested on five different ILSVRC2012 \emph{val} \cite{ILSVRC15} splits (each containing $10K$ images) and specified unknowns. OSR is performed on extractions from various pre-trained architectures. COSTARR (ours) is the final method in each network series and the best scores for each respective architecture and unknowns dataset are in \textbf{bold}.}
\end{table}

\begin{figure*}[t!]
    \centering
    \subfloat[a][iNaturalist]{
        \includegraphics[width=0.48\textwidth]{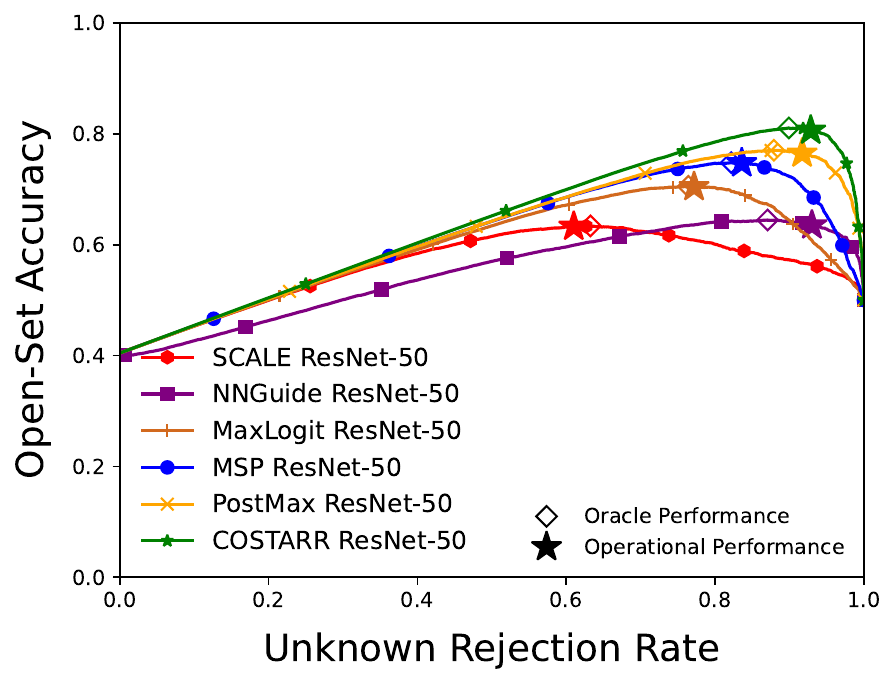}
        \label{fig:res-osa-inat}} 
    \subfloat[b][NINCO]{
        \includegraphics[width=0.48\textwidth]{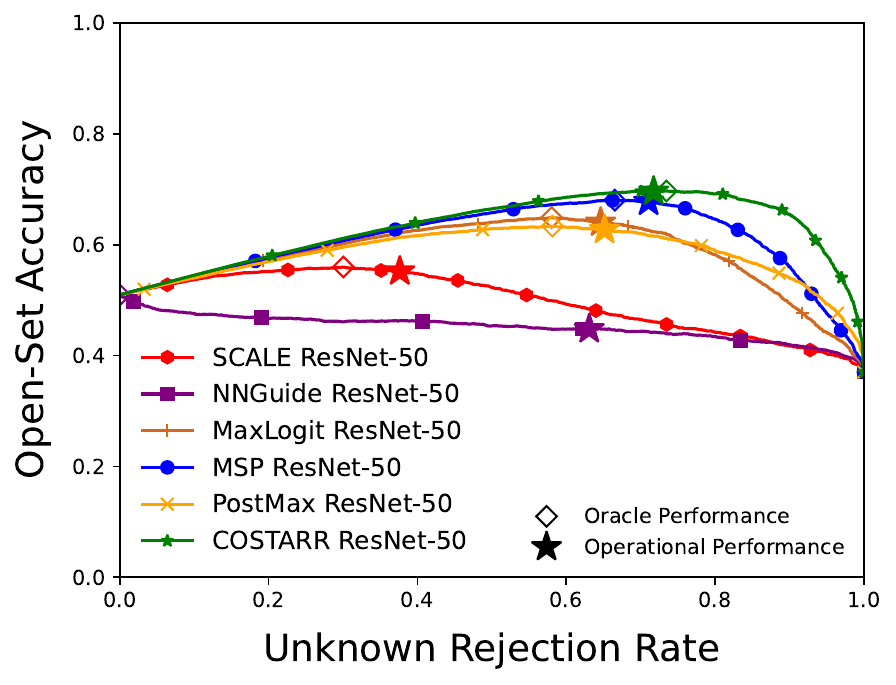}
        \label{fig:res-osa-ninco}} \\
    \subfloat[a][OpenImage-O]{
        \includegraphics[width=0.48\textwidth]{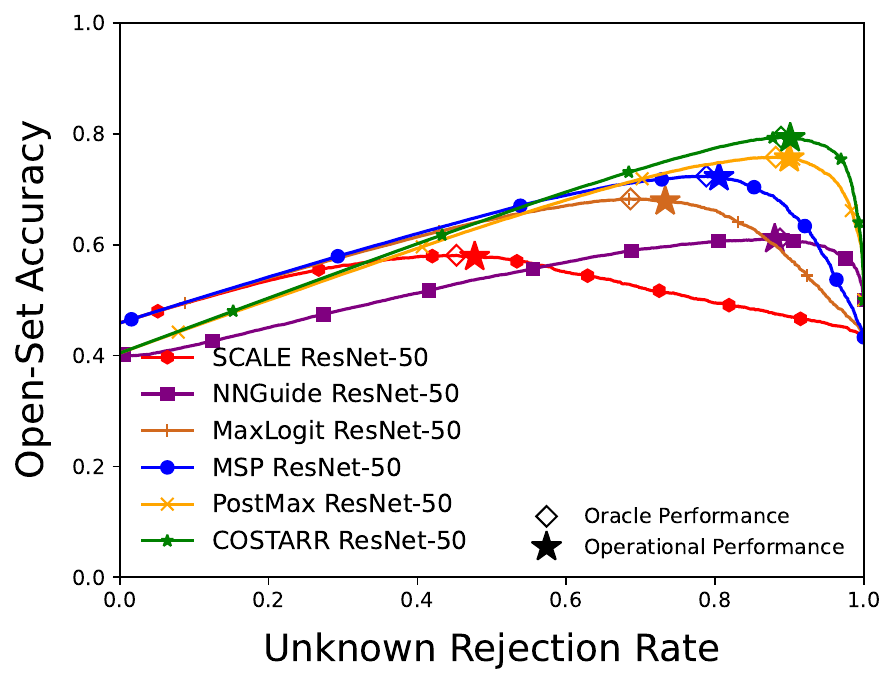}
        \label{fig:res-osa-open}} 
    \subfloat[b][Textures]{
        \includegraphics[width=0.48\textwidth]{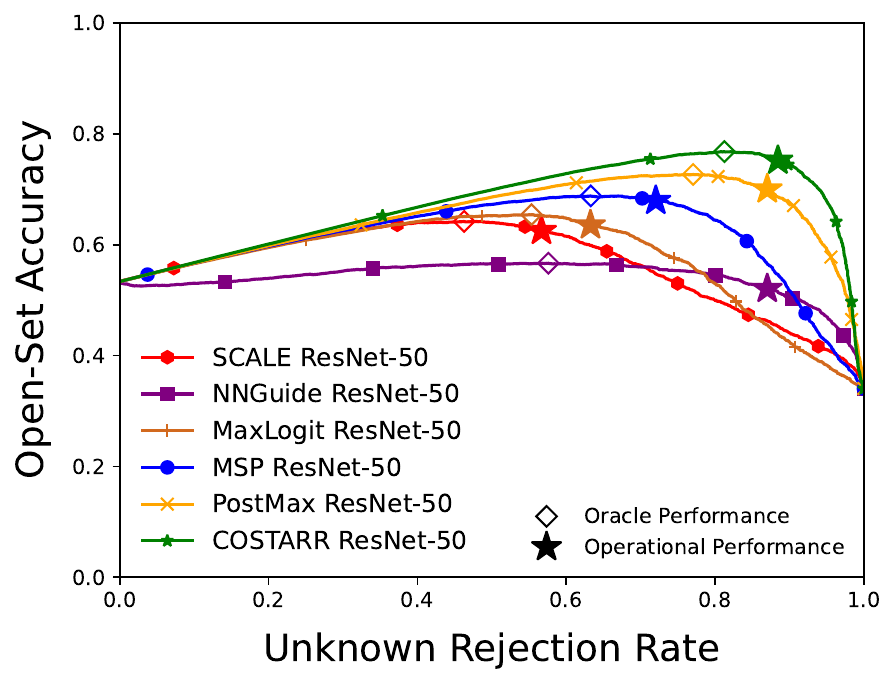}
        \label{fig:res-osa-text}} \\
    \Caption[fig:res-osa]{Open-Set Accuracy Curves}{The OSA curves of all methods for ResNet-50 (same experimental setup as \tab{oosa}). A $\star$ signifies the peak performance (OOSA) of each method.}
\end{figure*}

\begin{figure*}[t!]
    \centering
    \subfloat[a][iNaturalist]{
        \includegraphics[width=0.48\textwidth]{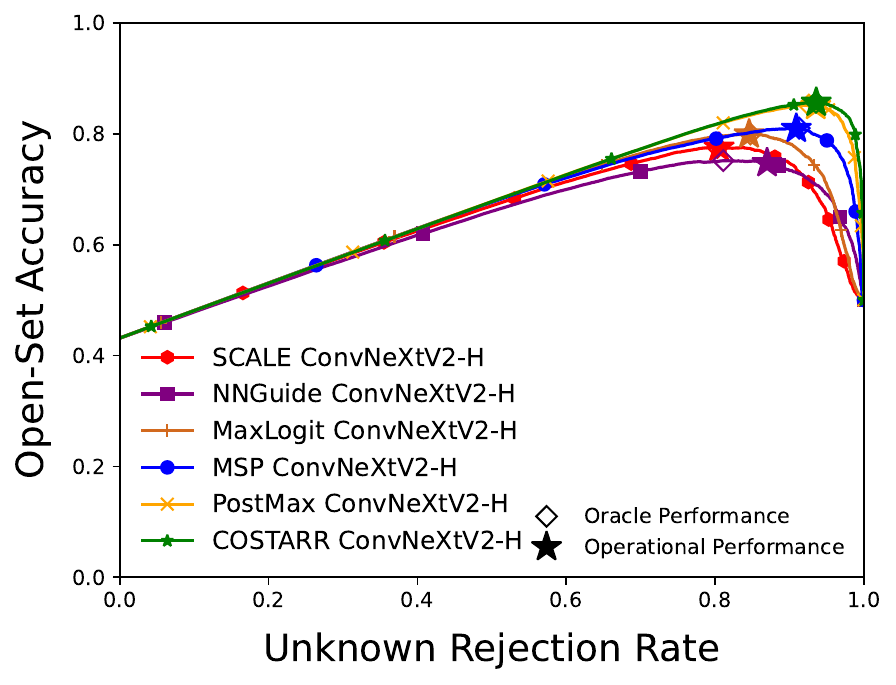}
        \label{fig:conv-osa-inat}} 
    \subfloat[b][NINCO]{
        \includegraphics[width=0.48\textwidth]{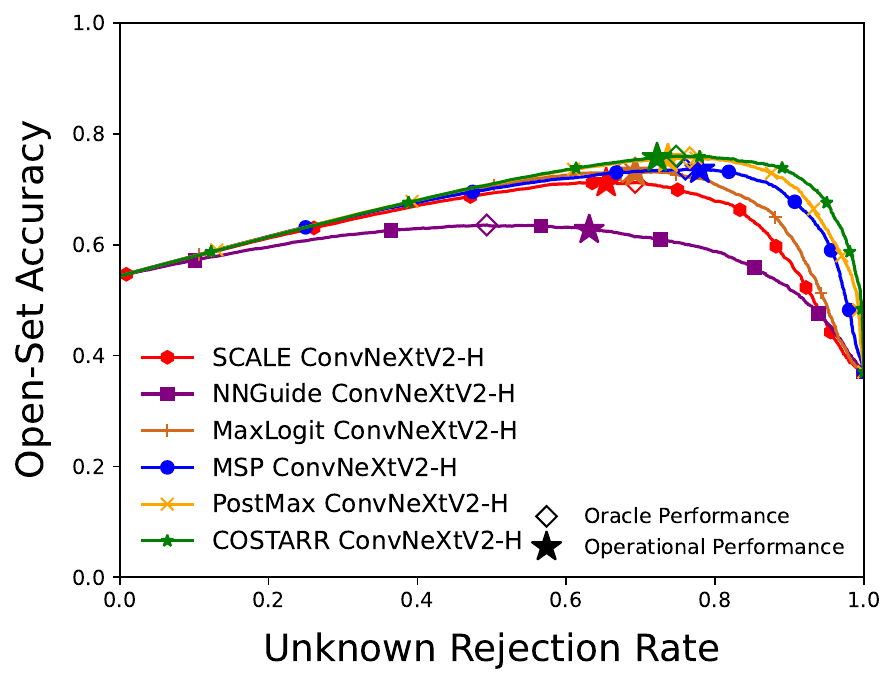}
        \label{fig:conv-osa-ninco}} \\
    \subfloat[a][OpenImage-O]{
        \includegraphics[width=0.48\textwidth]{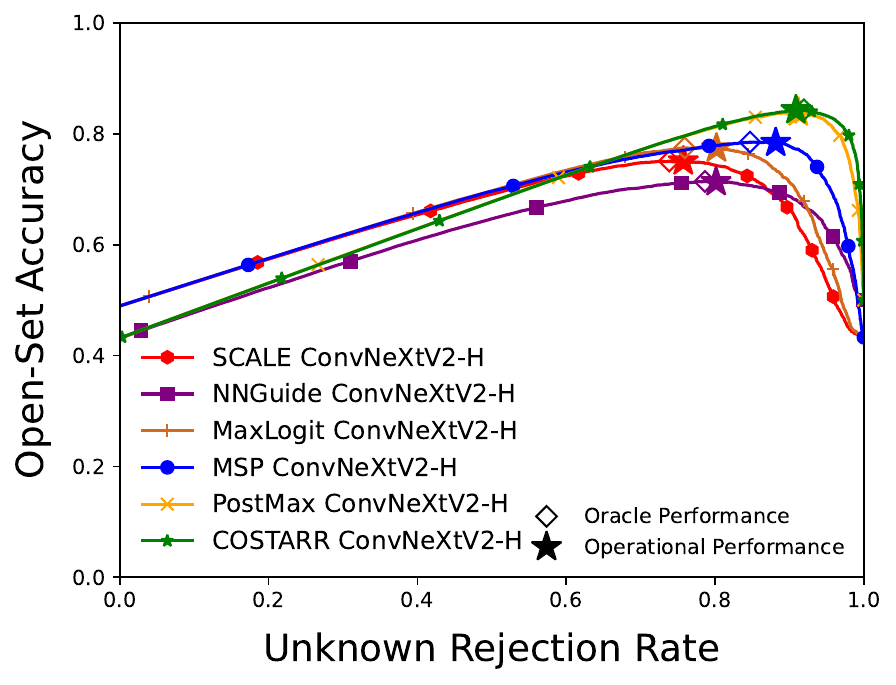}
        \label{fig:conv-osa-open}} 
    \subfloat[b][Textures]{
        \includegraphics[width=0.48\textwidth]{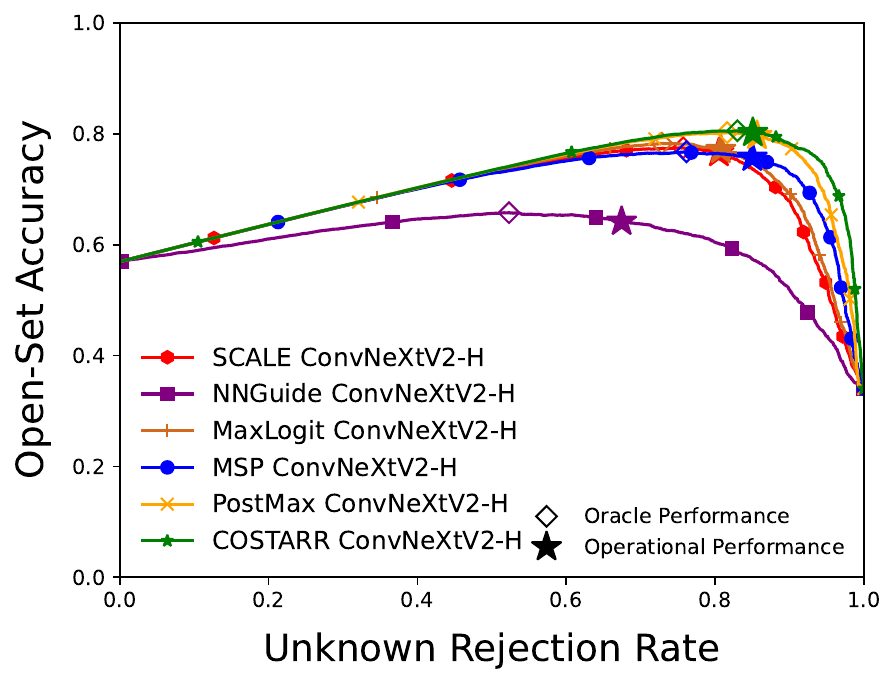}
        \label{fig:conv-osa-text}} \\
    \Caption[fig:conv-osa]{Open-Set Accuracy Curves}{The OSA curves of all methods for ConvNeXtV2-H (same experimental setup as \tab{oosa}). A $\star$ signifies the peak performance (OOSA) of each method.}
\end{figure*}

\begin{figure*}[t!]
    \centering
    \subfloat[a][iNaturalist]{
        \includegraphics[width=0.48\textwidth]{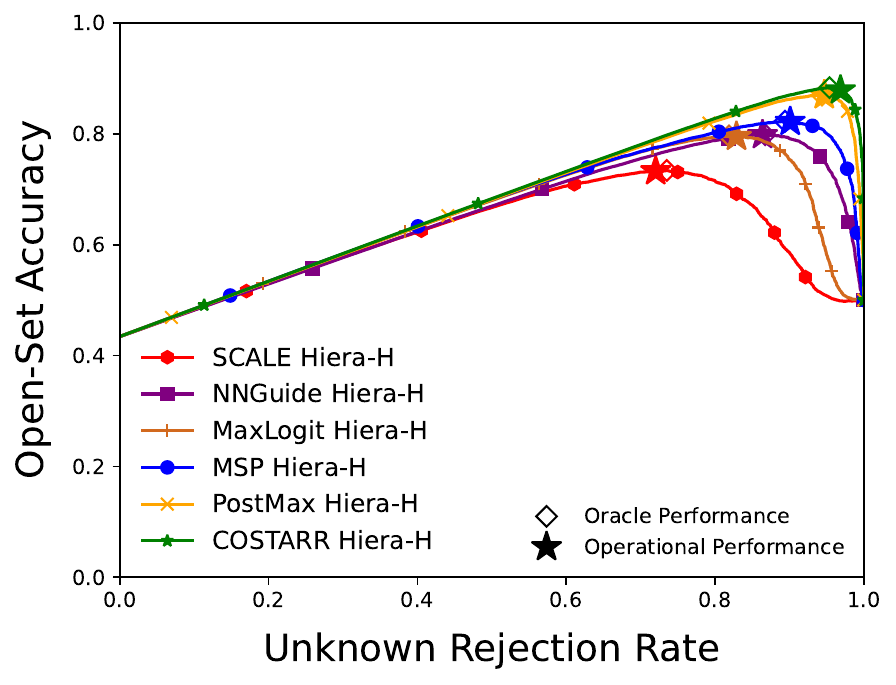}
        \label{fig:hiera-osa-inat}} 
    \subfloat[b][NINCO]{
        \includegraphics[width=0.48\textwidth]{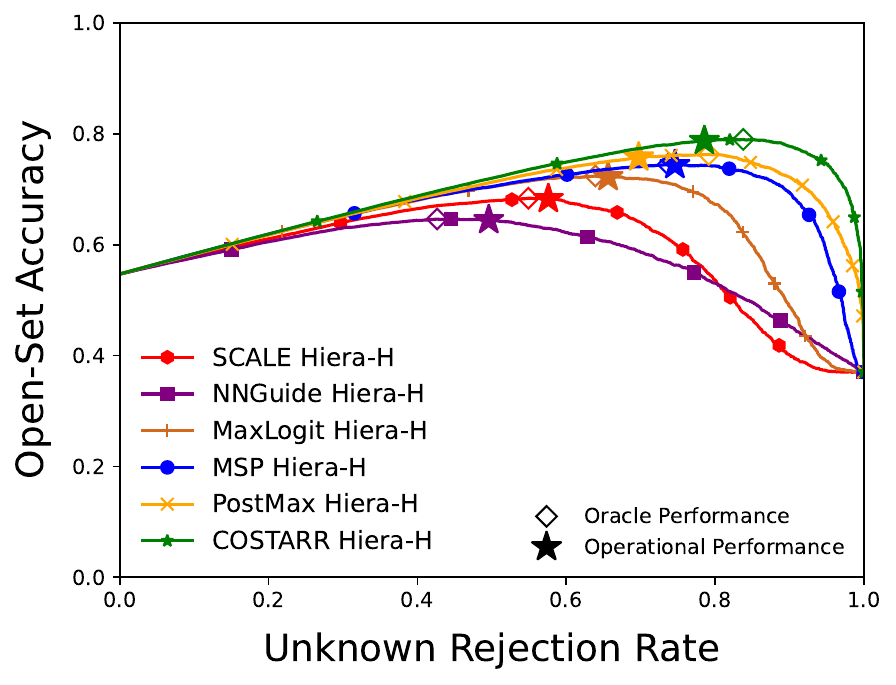}
        \label{fig:hiera-osa-ninco}} \\
    \subfloat[a][OpenImage-O]{
        \includegraphics[width=0.48\textwidth]{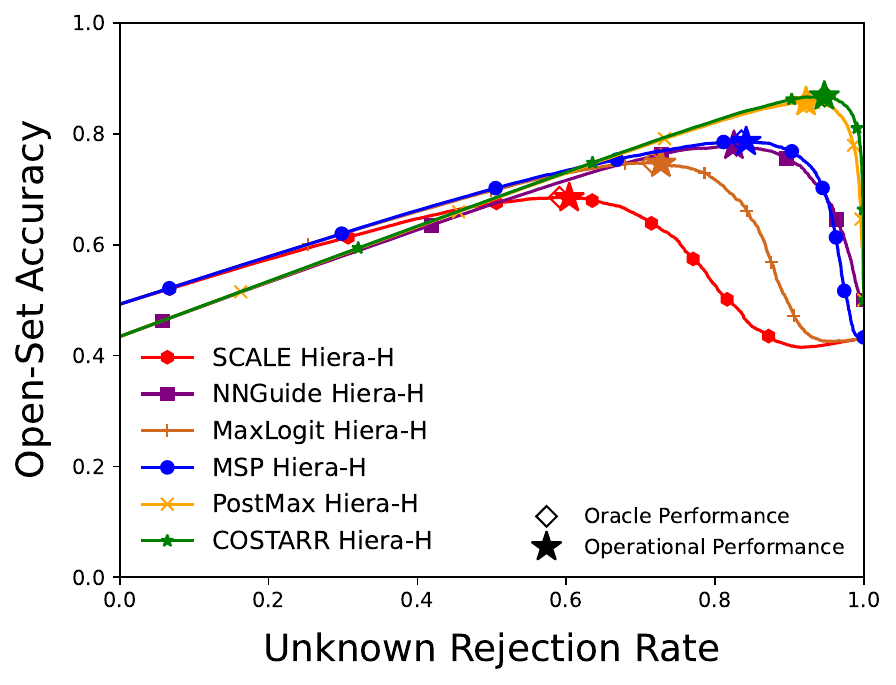}
        \label{fig:hiera-osa-open}} 
    \subfloat[b][Textures]{
        \includegraphics[width=0.48\textwidth]{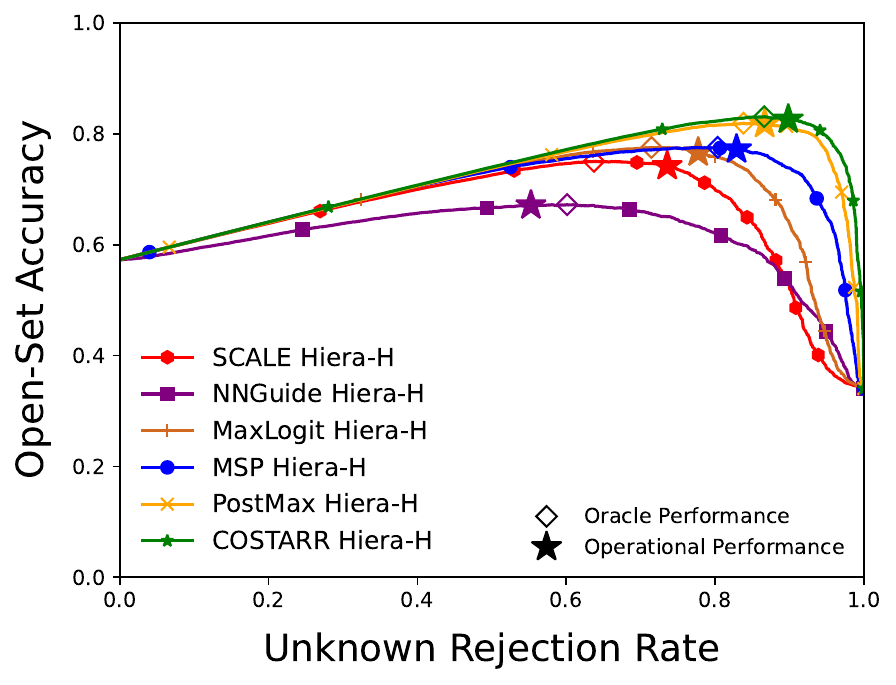}
        \label{fig:hiera-osa-text}} \\
    \Caption[fig:hiera-osa]{Open-Set Accuracy Curves}{The OSA curves of all methods for Hiera-H (same experimental setup as \tab{oosa}). A $\star$ signifies the peak performance (OOSA) of each method.}
\end{figure*}

\begin{figure*}[t!]
    \centering
    \subfloat[a][iNaturalist]{
        \includegraphics[width=0.48\textwidth]{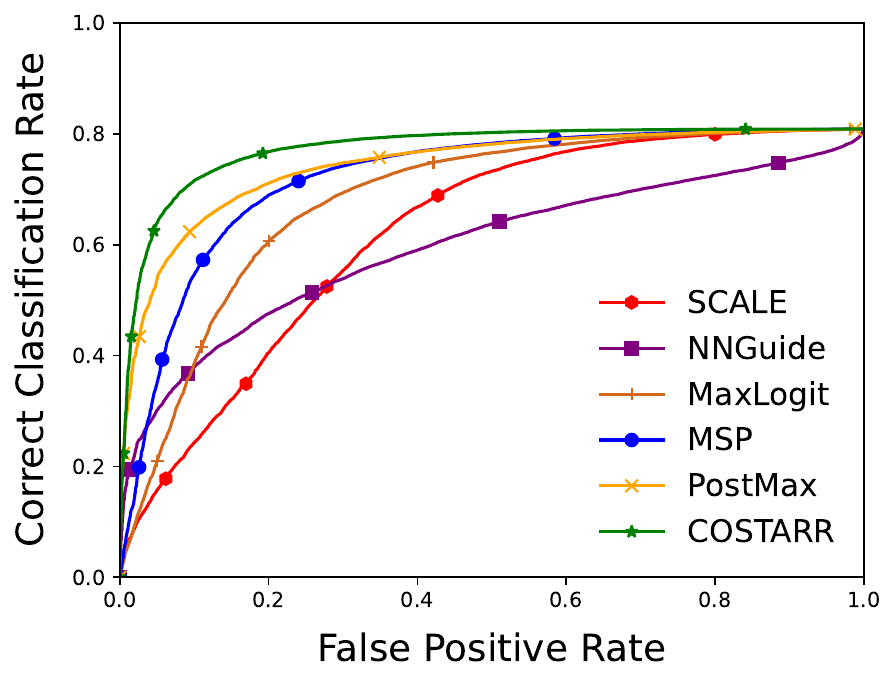}
        \label{fig:res-oscr-inat}} 
    \subfloat[b][NINCO]{
        \includegraphics[width=0.48\textwidth]{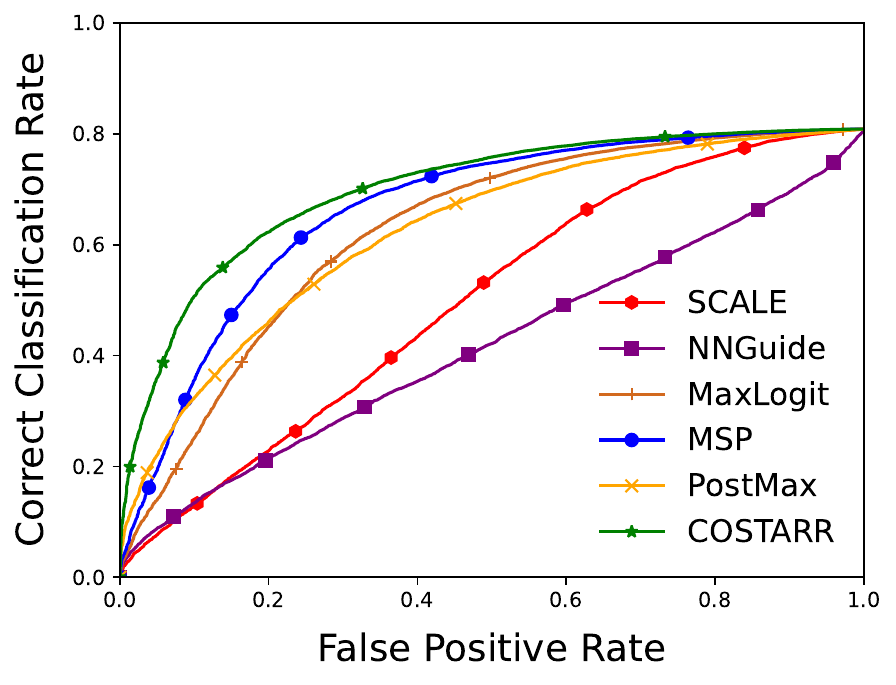}
        \label{fig:res-oscr-ninco}} \\
    \subfloat[a][OpenImage-O]{
        \includegraphics[width=0.48\textwidth]{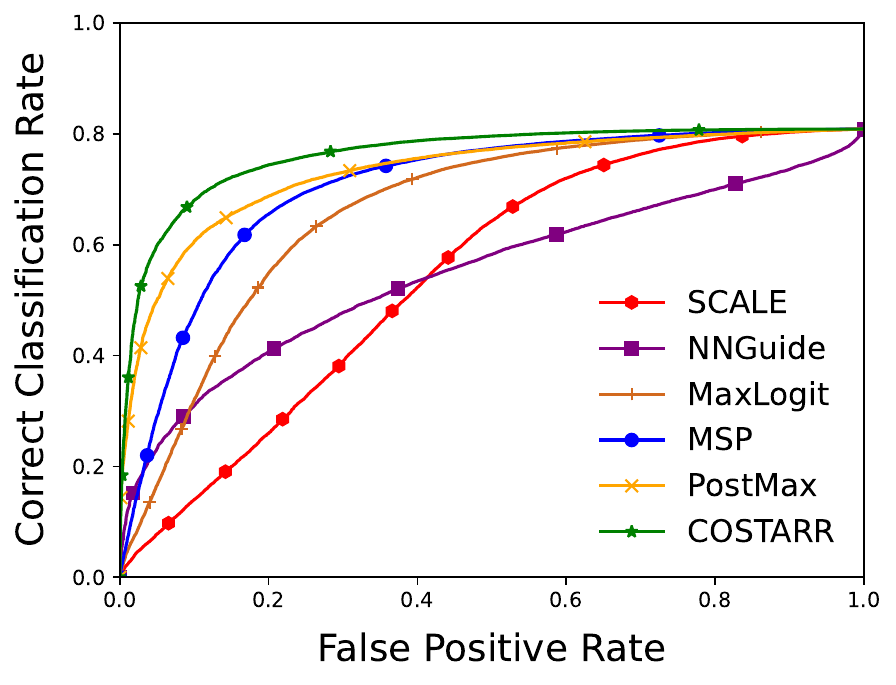}
        \label{fig:res-oscr-open}} 
    \subfloat[b][Textures]{
        \includegraphics[width=0.48\textwidth]{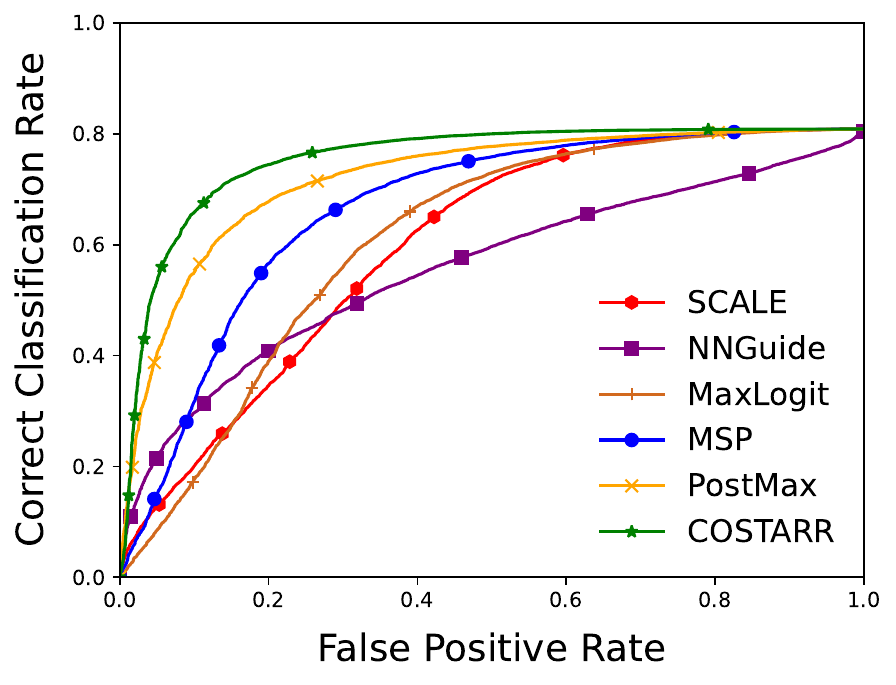}
        \label{fig:res-oscr-text}} \\
    \Caption[fig:res-oscr]{Open-Set Classification Rate Curves}{The OSCR curves of all methods for ResNet-50 (same experimental setup as \tab{auroc}).}
\end{figure*}

\begin{figure*}[t!]
    \centering
    \subfloat[a][iNaturalist]{
        \includegraphics[width=0.48\textwidth]{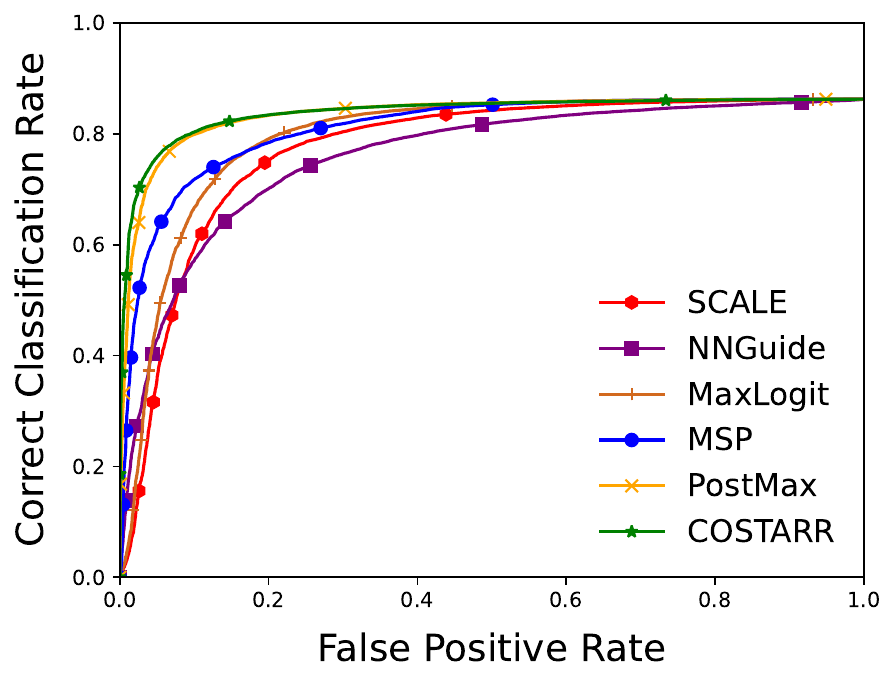}
        \label{fig:conv-oscr-inat}} 
    \subfloat[b][NINCO]{
        \includegraphics[width=0.48\textwidth]{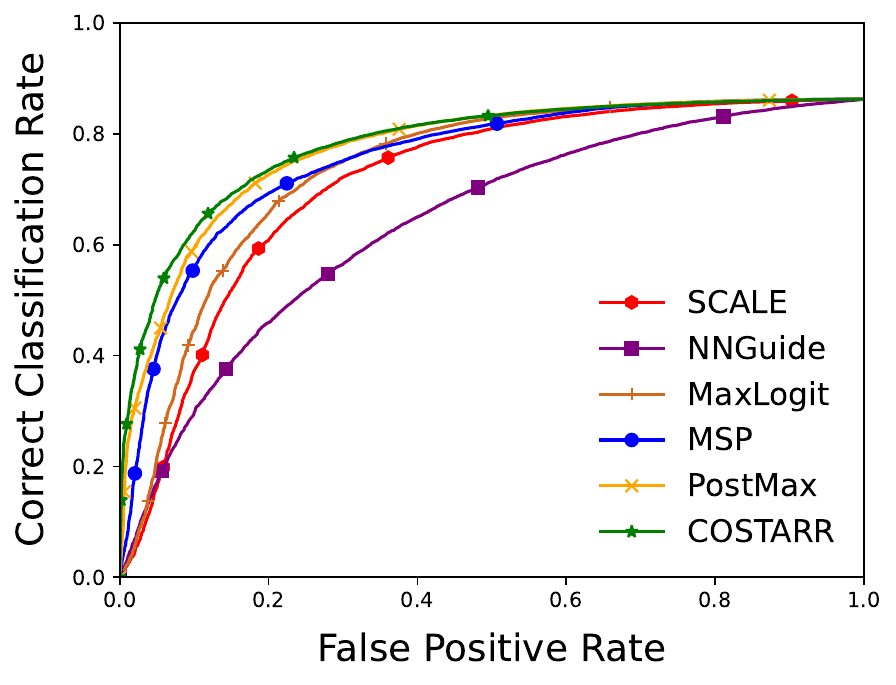}
        \label{fig:conv-oscr-ninco}} \\
    \subfloat[a][OpenImage-O]{
        \includegraphics[width=0.48\textwidth]{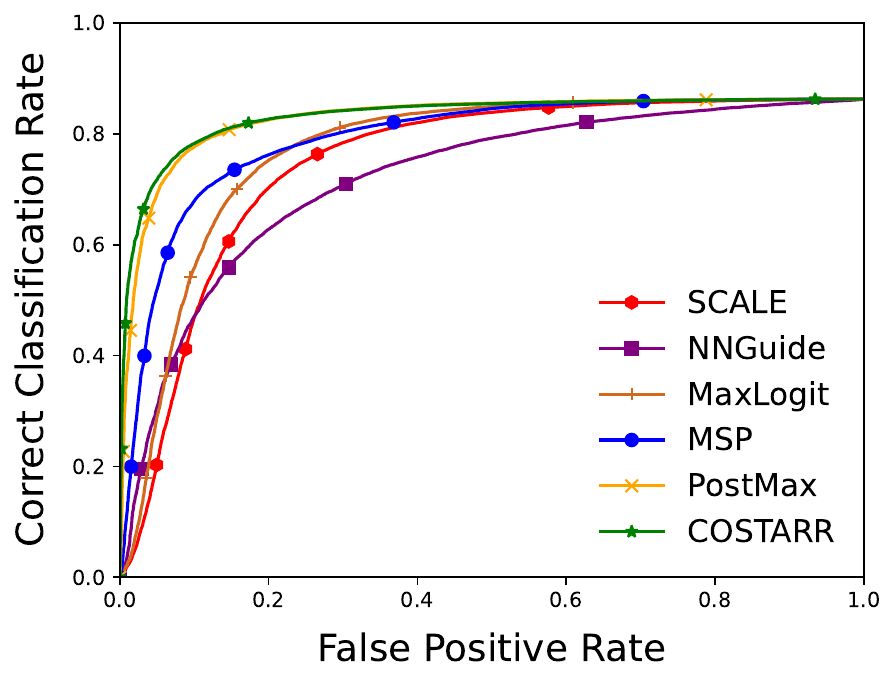}
        \label{fig:conv-oscr-open}} 
    \subfloat[b][Textures]{
        \includegraphics[width=0.48\textwidth]{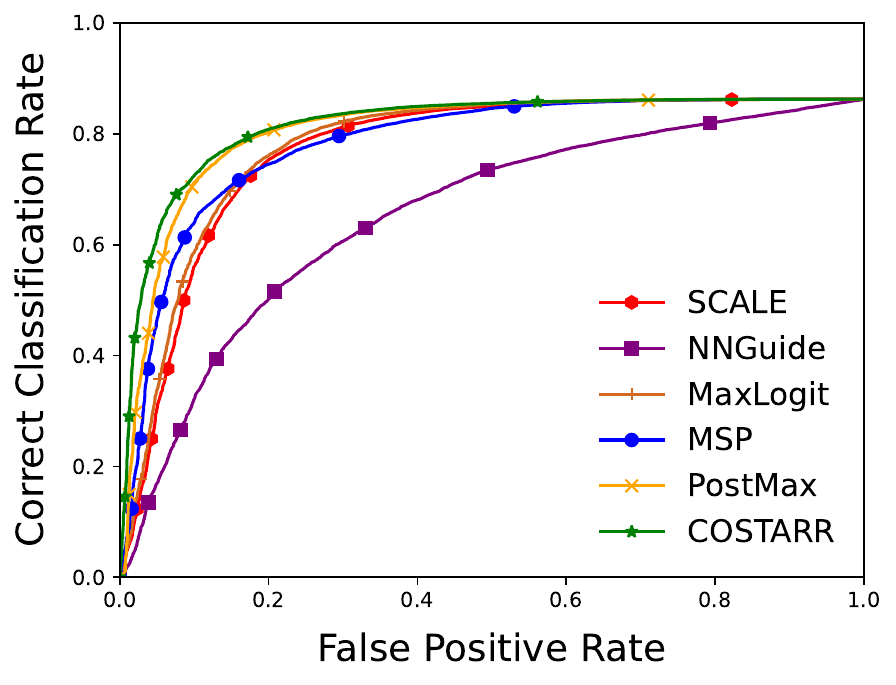}
        \label{fig:conv-oscr-text}} \\
    \Caption[fig:conv-oscr]{Open-Set Classification Rate Curves}{The OSCR curves of all methods for ConvNeXtV2-H (same experimental setup as \tab{auroc}).}
\end{figure*}

\begin{figure*}[t!]
    \centering
    \subfloat[a][iNaturalist]{
        \includegraphics[width=0.48\textwidth]{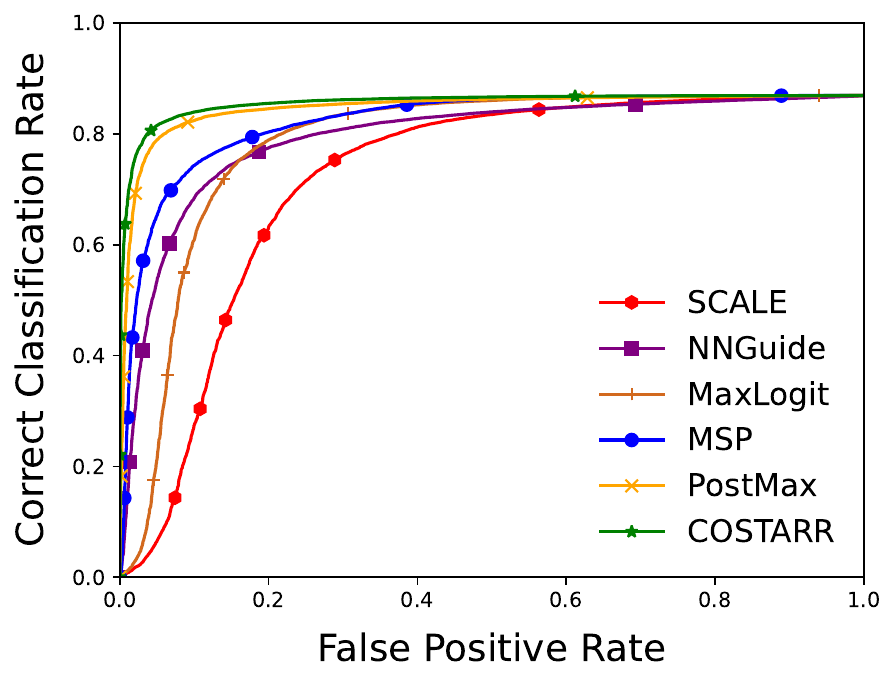}
        \label{fig:hiera-oscr-inat}}
    \subfloat[b][NINCO]{
        \includegraphics[width=0.48\textwidth]{Hiera-H_OSCR_NINCO.pdf}
        \label{fig:hiera-oscr-ninco}} \\
    \subfloat[a][OpenImage-O]{
        \includegraphics[width=0.48\textwidth]{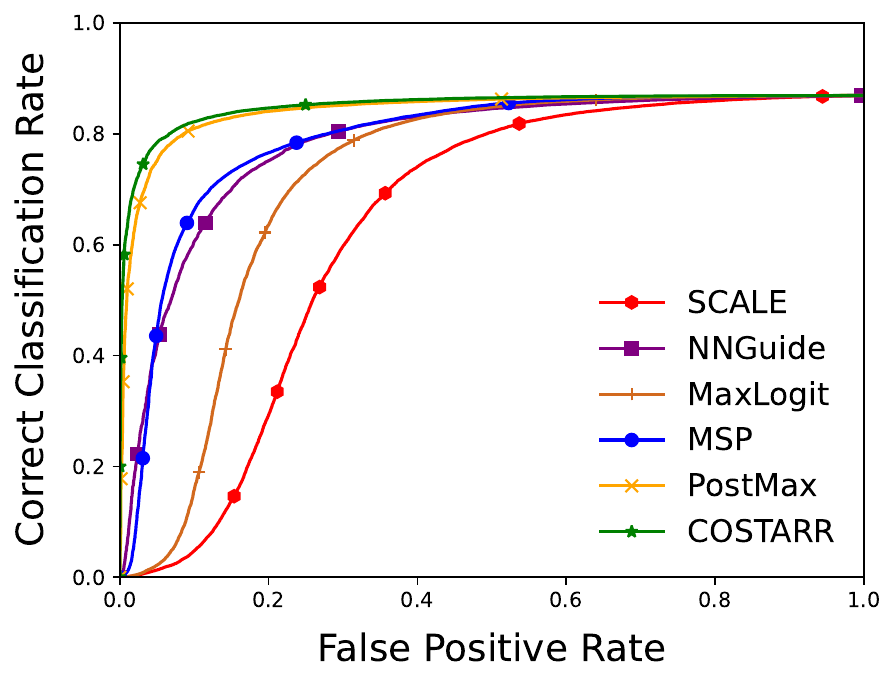}
        \label{fig:hiera-oscr-open}} 
    \subfloat[b][Textures]{
        \includegraphics[width=0.48\textwidth]{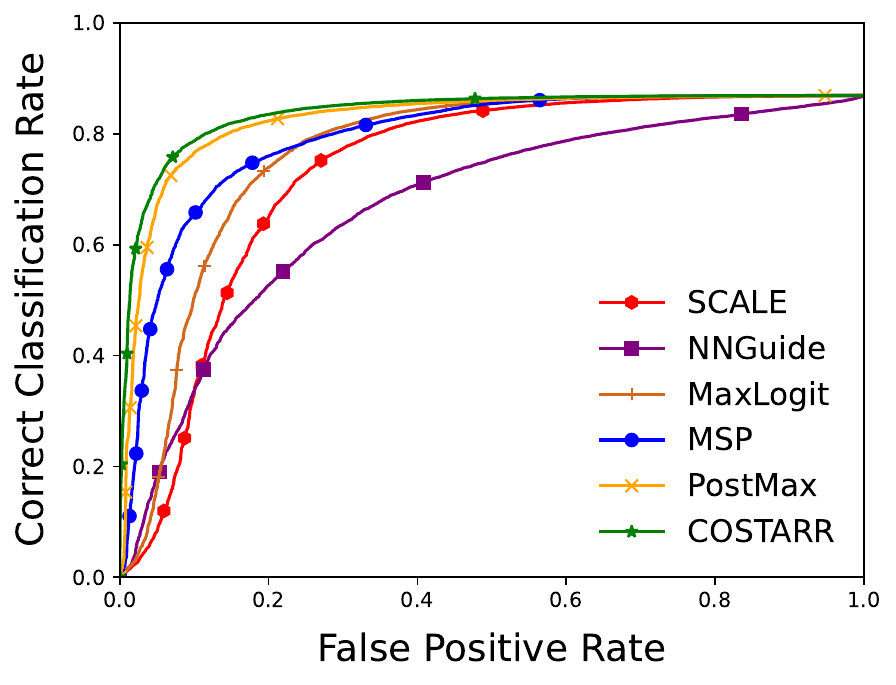}
        \label{fig:hiera-oscr-text}} \\
    \Caption[fig:hiera-oscr]{Open-Set Classification Rate Curves}{The OSCR curves of all methods for Hiera-H (same experimental setup as \tab{auroc}).}
\end{figure*}

\begin{table}[t!]
\begin{tabular}{c?c?cccc}
\multicolumn{1}{c?}{\rotatebox[origin=c]{90}{{\textbf{Arch}}}} & \multicolumn{1}{c?}{\textbf{Method}} & \textbf{iNat} & \textbf{NINCO} & \textbf{Open-O} & \textbf{Text} \\ \noalign{\hrule height 1.0pt}
\multirow{7}{*}{\rotatebox[origin=c]{90}{{ResNet-50}}} & SCALE & 0.756 & 0.604 & 0.663 & 0.729 \\
 & NNGuide & 0.715 & 0.492 & 0.656 & 0.665 \\
 & MaxLogit & 0.804 & 0.729 & 0.777 & 0.714 \\
 & COMBOOD$\dagger$ & 0.871 & N/A & 0.866 & \textbf{0.970} \\
 & MSP & 0.848 & 0.768 & 0.823 & 0.774 \\
 & PostMax & 0.882 & 0.722 & 0.862 & 0.852 \\
 & COSTARR & \textbf{0.931} & \textbf{0.816} & \textbf{0.910} & 0.905 \\ \hline
\multirow{6}{*}{\rotatebox[origin=c]{90}{{ConvNeXt-L}}} & SCALE & 0.773 & 0.721 & 0.734 & 0.76 \\
 & NNGuide & 0.676 & 0.52 & 0.704 & 0.507 \\
 & MaxLogit & 0.836 & 0.765 & 0.794 & 0.792 \\
 & MSP & 0.876 & 0.799 & 0.848 & 0.823 \\
 & PostMax & 0.920 & 0.817 & 0.911 & 0.880 \\
 & COSTARR & \textbf{0.944} & \textbf{0.863} & \textbf{0.934} & \textbf{0.912} \\ \hline
 \multirow{6}{*}{\rotatebox[origin=c]{90}{{ConvNeXtV2-H}}} & SCALE & 0.878 & 0.803 & 0.849 & 0.879 \\
 & NNGuide & 0.860 & 0.715 & 0.811 & 0.731 \\
 & MaxLogit & 0.899 & 0.820 & 0.869 & 0.881 \\
 & MSP & 0.908 & 0.831 & 0.883 & 0.873 \\
 & PostMax & 0.949 & 0.864 & 0.940 & 0.912 \\
 & COSTARR & \textbf{0.954} & \textbf{0.875} & \textbf{0.946} & \textbf{0.925} \\ \hline
\multirow{6}{*}{\rotatebox[origin=c]{90}{{ViT-H}}} & SCALE & 0.847 & 0.742 & 0.767 & 0.842 \\
 & NNGuide & 0.728 & 0.525 & 0.758 & 0.493 \\
 & MaxLogit & 0.889 & 0.786 & 0.817 & 0.858 \\
 & MSP & 0.910 & 0.828 & 0.869 & 0.867 \\
 & PostMax & 0.957 & 0.857 & 0.947 & 0.925 \\
 & COSTARR & \textbf{0.970} & \textbf{0.896} & \textbf{0.959} & \textbf{0.942} \\ \hline
\multirow{6}{*}{\rotatebox[origin=c]{90}{{Hiera-H}}} & SCALE & 0.814 & 0.725 & 0.716 & 0.829 \\
 & NNGuide & 0.896 & 0.690 & 0.881 & 0.744 \\
 & MaxLogit & 0.883 & 0.786 & 0.802 & 0.860 \\
 & MSP & 0.920 & 0.835 & 0.875 & 0.881 \\
 & PostMax & 0.959 & 0.872 & 0.953 & 0.930 \\
 & COSTARR & \textbf{0.973} & \textbf{0.907} & \textbf{0.963} & \textbf{0.947}
\end{tabular}
\Caption[tab:auroc]{Area Under Receiver Operating Characteristic Curve}{The AUROC ($\uparrow$) of all methods. To compute, we tested methods using ILSVRC2012 \emph{val} \cite{ILSVRC15} ($50K$ images) as knowns and specified unknowns. COSTARR (ours) is the final method in each network series and the best scores for each respective architecture and unknowns dataset are in \textbf{bold}. $\dagger$ indicates a result has been transposed from \cite{rajasekaran2024combood}.}
\end{table}

\section{Different OOSA Validation}
For consistency of comparison with prior work, we also report performance on the same validation from Cruz \etal{cruz2025operational} in \tab{oosa-hard}: ImageNetV2 (10K images) as knowns and 21K-P Hard (9.8K images) as unknowns with their "contaiminated" validation process. 
Similar to the results in \tab{oosa}, COSTARR outperforms all methods across all architectures except ViT-H.

\clearpage
\section{Statistical Testing}
\label{sec:allstats}

In Tables \ref{tab:oo-clean-stat}, \ref{tab:oo-dirty-stat}, \ref{tab:hard-clean-stat}, and \ref{tab:hard-dirty-stat} we present P-values from statistical testing of COSTARR vs different algorithms on different architectures and datasets for validation and testing. 
COSTARR is statistically significantly better in almost all cases and is never statistically worse. 

While \cite{cruz2025operational} introduce splits and a testing procedure, we modified that process somewhat for this paper.  For the splits for the knowns, there are non-overlapping samples from each class, and a set of unknowns varies by dataset, so they assumed independence and used two-sided paired t-tests.   But when combining data over different architectures or different unknown sets, the independence breaks down. Thus we use Wilcoxon signed rank test, as implemented in scipy,  with Bonferroni correction to account for the different number of tests being considered.  

We obtained the splits from them and recomputed statistical significance using Wilcoxon signed rank test, as implemented in scipy,  with Bonferroni correction. The results do not significantly change the conclusions from their paper, but we could only answer that by doing the proper non-parametric test.  This also sets up the work so that future tests can properly compare with COSTARR.  The splits will be included in with released code. 

\begin{table*}[!ht]
    \centering
   \begin{tabular}{|l|l|l|l|l|l|}
    \hline
        \textit{Net  /  Alg}  & \textbf{PostMax} & \textbf{MSP} & \textbf{MaxLogit} & \textbf{NNGuide} & \textbf{SCALE} \\ \hline
        Reset-50 P-Value (N=15) & 1.82E-08 & 1.16E-09 & 7.93E-12 & 3.87E-12 & 2.88E-15 \\ \hline
        ConvNext-L P-Value (N=15)& 1.31E-07 & 1.24E-11 & 1.82E-10 & 5.70E-17 & 2.11E-11 \\ \hline
        ConvNextV2-H P-Value (N=15)& 5.23E-10 & 1.79E-09 & 5.03E-09 & 1.73E-15 & 2.08E-10 \\ \hline
        ViT-H  P-Value (N=15)& 8.81E-06 & 1.09E-10 & 9.13E-11 & 2.15E-14 & 5.99E-12 \\ \hline
        Hiera-H  P-Value (N=15)& 9.24E-05 & 2.24E-11 & 8.01E-11 & 1.45E-09 & 2.26E-11 \\ \hline \hline
        \textbf{Overall P-Value (N=75)} & 9.35E-12 & 1.78E-37 & 3.02E-39 & 1.48E-30 & 5.28E-30 \\ \hline
    \end{tabular}
    \Caption[tab:oo-clean-stat]{Statistics comparing OOSA performance of COSTARR vs. different algorithms on "clean data" where  Open-Images is a surrogate set}{P-values of the null hypothesis: that the mean performance is the same as COSTARR. This uses a subset of Open-Images as the surrogate set to select the threshold for the given network. The statistics corresponding to each architecture use five different splits of the validation data and unknowns drawn from the full set of iNaturalist, NINCO, and OpenImage\_O. There are N=15 runs per architecture and 75 runs overall. Since OOSA computes thresholds directly on the surrogate set, there are {\em NO free/tuned parameters} in these experiments for any architecture. This is computed by Wilcoxon signed rank test with Bonferroni correction.  All tests are statistically very significant. 
  }
\end{table*}

\begin{table*}[!ht]

    \centering
    \begin{tabular}{|l|l|l|l|l|l|}
    \hline
        \textit{Net  /  Alg}  & \textbf{PostMax} & \textbf{MSP} & \textbf{MaxLogit} & \textbf{NNGuide} & \textbf{SCALE} \\ \hline
        {Reset-50 P-value  (N=40)}   & 4.163E-08 & 3.067E-11 & 2.341E-21 & 3.955E-11 & 3.953E-21 \\ \hline
        {ConvNext-L P-value (N=40)}   & 6.710E-05 & 6.181E-08 & 4.354E-20 & 8.862E-15 & 7.392E-23 \\ \hline
        {ConvNextV2-H P-value (N=40)}   & 1.433E-03 & \textcolor{purple}{2.152E-01} & 3.098E-15 & 3.385E-09 & 1.910E-18 \\ \hline
        {ViT-H  P-value (N=40) }   & 1.006E-04 & 2.967E-07 & 3.704E-19 & 2.368E-18 & 6.521E-24 \\ \hline
        {Hiera-H P-value (N=40)}   & 1.372E-04 & 4.188E-11 & 5.033E-23 & 1.849E-13 & 1.549E-24 \\ \hline \hline
        \textbf{Overall P-value (N=200})  &  9.354-12 & 1.784-37 & 3.025E-39 & 1.484E-30 & 5.282E-30  \\ \hline
    \end{tabular}
\Caption[tab:oo-dirty-stat]{Statistics comparing OOSA performance of COSTARR vs. different algorithms on "Contaminated data" where  Open-Images is Surrogate set}{P-values of null hypothesis: that the mean performance is the same as COSTARR. This uses a subset of Open-Images as the surrogate set to select the threshold for the given network. The statistics corresponding to each architecture use five different splits of the validation data and unknowns drawn from the full set of iNaturalist, NINCO, OpenImage\_O, Places, SUN, Textures, easy\_i21k, and hard\_i21k, so it includes results where many corrupted/overlapping classes/images, with N=40 runs per architecture and 200 overall. Since OOSA computes thresholds directly on the surrogate set, there are {\em NO free/tuned parameters} in these experiments for any architecture. This is computed by Wilcoxon signed rank test with Bonferroni correction.  All tests except ConvNextV2-H for MSP (shown in \textcolor{purple}{purple}) are statistically very significant. 
}
\end{table*}

\begin{table*}[!ht]
    \centering
    \begin{tabular}{|l|l|l|l|l|l|}
    \hline
        Net/Alg & PostMax & MSP & MaxLogit & NNGuide & SCALE \\ \hline
        Reset-50 P-value (N=15)& 1.25E-11 & 7.72E-09 & 1.28E-09 & 3.83E-13 & 6.07E-13 \\ \hline
        ConvNext-L P-value (N=15) & 3.95E-08 & 1.58E-08 & 2.36E-09 & 4.11E-13 & 1.19E-10 \\ \hline
        ConvNextV2-H P-value (N=15) & 9.43E-08 & 4.15E-08 & 2.15E-08 & 4.23E-10 & 9.27E-09 \\ \hline
        ViT-H  P-value (N=15)& 4.34E-08 & 8.18E-09 & 1.41E-09 & 1.66E-11 & 1.08E-10 \\ \hline
        Hiera-H P-value (N=15) & 3.60E-08 & 1.43E-08 & 1.62E-09 & 1.19E-09 & 7.53E-11 \\ \hline\hline
        \textbf{Overall  P-value (N=75)}& 3.01E-06 & 6.41E-30 & 7.56E-33 & 7.95E-30 & 3.32E-32 \\ \hline
    \end{tabular}
\Caption[tab:hard-clean-stat]{Statistics comparing OOSA performance of COSTARR vs. different algorithms where  hard\_i21k is a Surrogate set with non-contaminated unknown sets}{P-values of the null hypothesis: that the mean performance is the same as COSTARR.  This uses a subset of hard\_i21k as the surrogate set to select the threshold for the given network. The statistics corresponding to each architecture use five different splits of the validation data and unknowns drawn from the non-contaminated set of iNaturalist, NINCO, and OpenImage\_O. It has N=15 runs per architecture and 75 overall.  This is computed by Wilcoxon signed rank test with Bonferroni correction.  All tests are statistically very significant. 
}
\end{table*}

\begin{table*}[!ht]
    \centering
    \begin{tabular}{|l|l|l|l|l|l|}
    \hline
        \textit{Net  /  Alg}  & \textbf{PostMax} & \textbf{MSP} & \textbf{MaxLogit} & \textbf{NNGuide} & \textbf{SCALE} \\ \hline
        Reset-50 P-Value (N=36) & 2.89E-12 & 4.27E-24 & 1.31E-30 & 3.11E-13 & 1.40E-17 \\ \hline
        ConvNext-L P-Value (N=36)& 4.18E-07 & 9.78E-18 & 3.43E-14 & 1.58E-13 & 2.36E-17 \\ \hline
        ConvNextV2-H P-Value (N=36)& 1.41E-02 & 1.59E-10 & 4.42E-09 & 1.43E-15 & 1.66E-09 \\ \hline
        ViT-H P-value (N=36) & \textcolor{purple}{7.77E-01} & 1.29E-15 & 1.33E-09 & 3.98E-16 & 1.56E-13 \\ \hline
        Hiera-H P-Value (N=36)& 4.28E-06 & 2.07E-19 & 6.17E-19 & 4.64E-14 & 3.91E-18 \\ \hline \hline
        \textbf{Overall  P-Value (N=176)}& 3.83E-10 & 4.85E-71 & 5.31E-60 & 1.50E-53 & 1.27E-60 \\ \hline 
    \end{tabular}
    \Caption[tab:hard-dirty-stat]{Statistics comparing OOSA performance of COSTARR vs. different algorithms where  hard\_i21k is a Surrogate set, similar to that used in PostMax Paper}{P-values of the null hypothesis that the mean performance is the same as COSTARR.  This uses a subset of hard\_i21k as the surrogate set to select the threshold for the given network. The statistics corresponding to each architecture use five different splits of the validation data and unknowns drawn from the full set of iNaturalist, NINCO, OpenImage\_O, Places, SUN, Textures, easy\_i21k,  so it includes results where many corrupted/overlapping classes/images, with N=36 runs per architecture and 176 overall, which is slightly less than when Open-Images is used for the surrogate set.     Since OOSA computes thresholds directly on the surrogate set, there are {\em NO free/tuned parameters} in these experiments for any architecture. This is computed by paired two-sided t-tests with Bonferroni correction.  All tests except  ViT-H for PostMax (shown in \textcolor{purple} purple) are statistically very significant. 
  }
\end{table*}

\clearpage

\section{Contamination}
Given the recent in-distribution contamination analysis performed by Bitterwolf \etal{bitterwolf2023or}, we excluded any datasets with $>$20\% contamination from the main paper.
However, to report performance on datasets used in prior SOTA evaluations \cite{cruz2025operational}, we show results in \tab{oosa-contam}.
Note, COSTARR models per-class confidence, so any significant overlap will hinder performance.
Nonetheless, overall performance is still statistically significant as shown in \sec{allstats}.

To analyze how in-distribution contaminated unknown datasets degrade evaluations, we examined the images responsible for the performance difference between PostMax and COSTARR for the Places and SUN datasets.
At the validation-predicted OOSA threshold, we looked at samples which PostMax said were unknown, but COSTARR labeled as known.
We compared these images with ImageNet training data from the closed-set predicted class.
From Places, we found every image within this performance gap has visually significant overlap with ImageNet training data, from SUN we found the same holds for nearly every image.
We include \emph{all} images from these examinations in Figures \ref{fig:contam-viz-1} \& \ref{fig:contam-viz-2} for Places and Figures \ref{fig:sun-contam-viz-1} and \ref{fig:sun-contam-viz-2}.
Labeling images as unknown while they are visually represented in the training set hinders OSR and OOD evaluations.
We can see from these figures, the images responsible for COSTARR's seemingly inferior performance to PostMax are actually knowns which have been mislabeled as unknowns.
Of course, all algorithms are affected by the presence of these mislabeled images.
When contaminated dataset is used as unknowns, even a perfect OSR system will have a reduced OOSA and AUROC score.

\begin{figure*}[t!]
  \centering
  \includegraphics[width=0.8\textwidth]{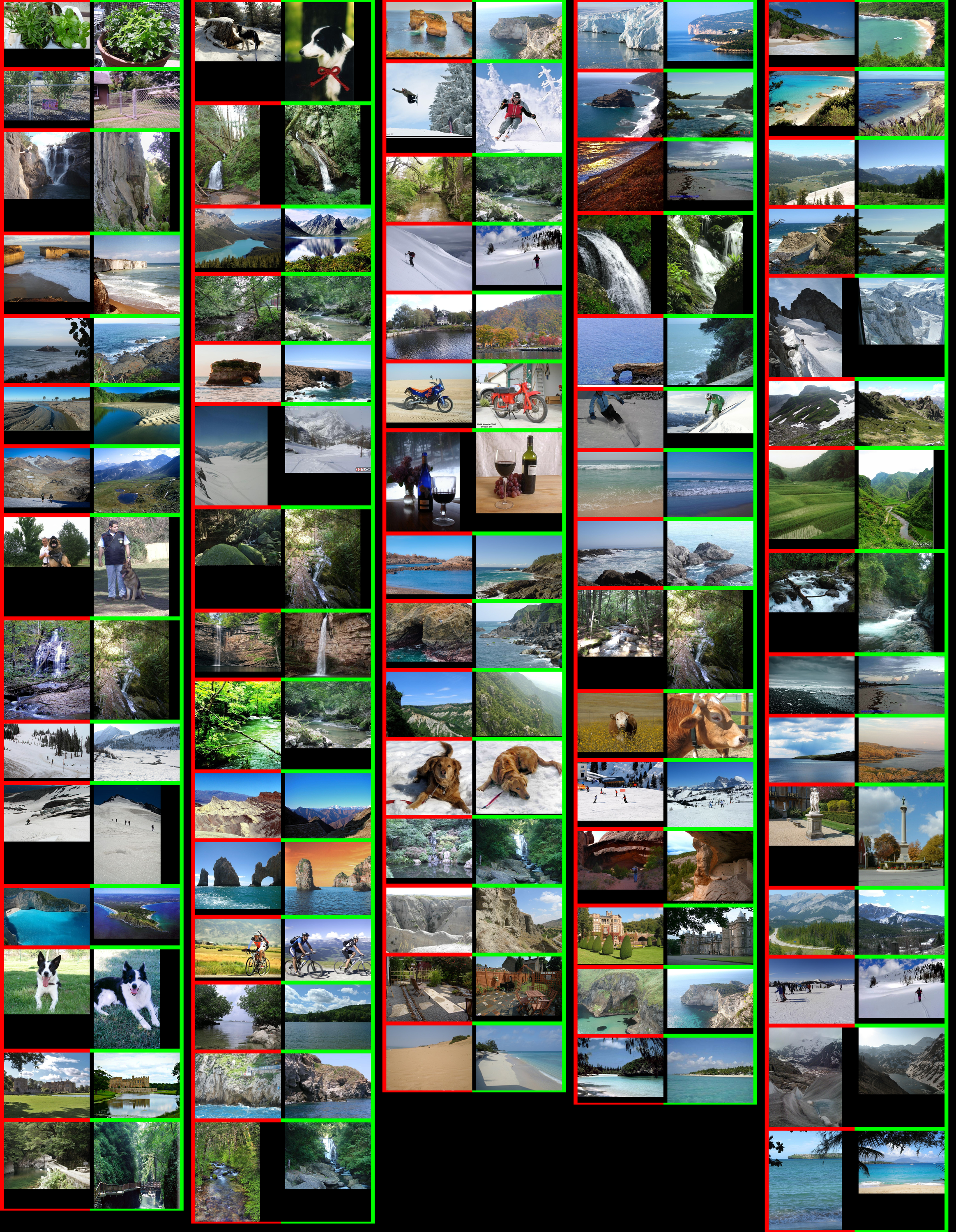}
\Caption[fig:contam-viz-1]{Data Contamination in Places}{We analyzed images from the OOD dataset Places \cite{zhou2017places} (red bordered images) compared with Imagenet training images (green bordered images). To select OOD images from Places\cite{zhou2017places}, we examined the validation threshold for PostMax and COSTARR on ViT-H, then selected those images from Places which PostMax would label unknown, but COSTARR would label known. Between this figure and Figure \ref{fig:contam-viz-2}, we present \emph{every image satisfying this constraint}, hence, these images are directly responsible for the performance difference between the algorithms (in terms of OOSA). To select ImageNet training images, we examined all images from the closed-set prediction class (the known class the network predicted the unknown image was) and selected the closest one. As all the images from Places (red bordered) are treated as unknowns and the networks has seen all of the ImageNet training data (green bordered), these falsely labeled unknowns are actually hindering the evaluation, given the clear overlap between known training data and supposed unknowns. The presence of these mislabeled unknowns is consistent with NINCO's \cite{bitterwolf2023or} observations, partially validating their claim of in-distribution dataset contamination.
}
\end{figure*}
\begin{figure*}[t!]
  \centering
  \includegraphics[width=0.8\textwidth]{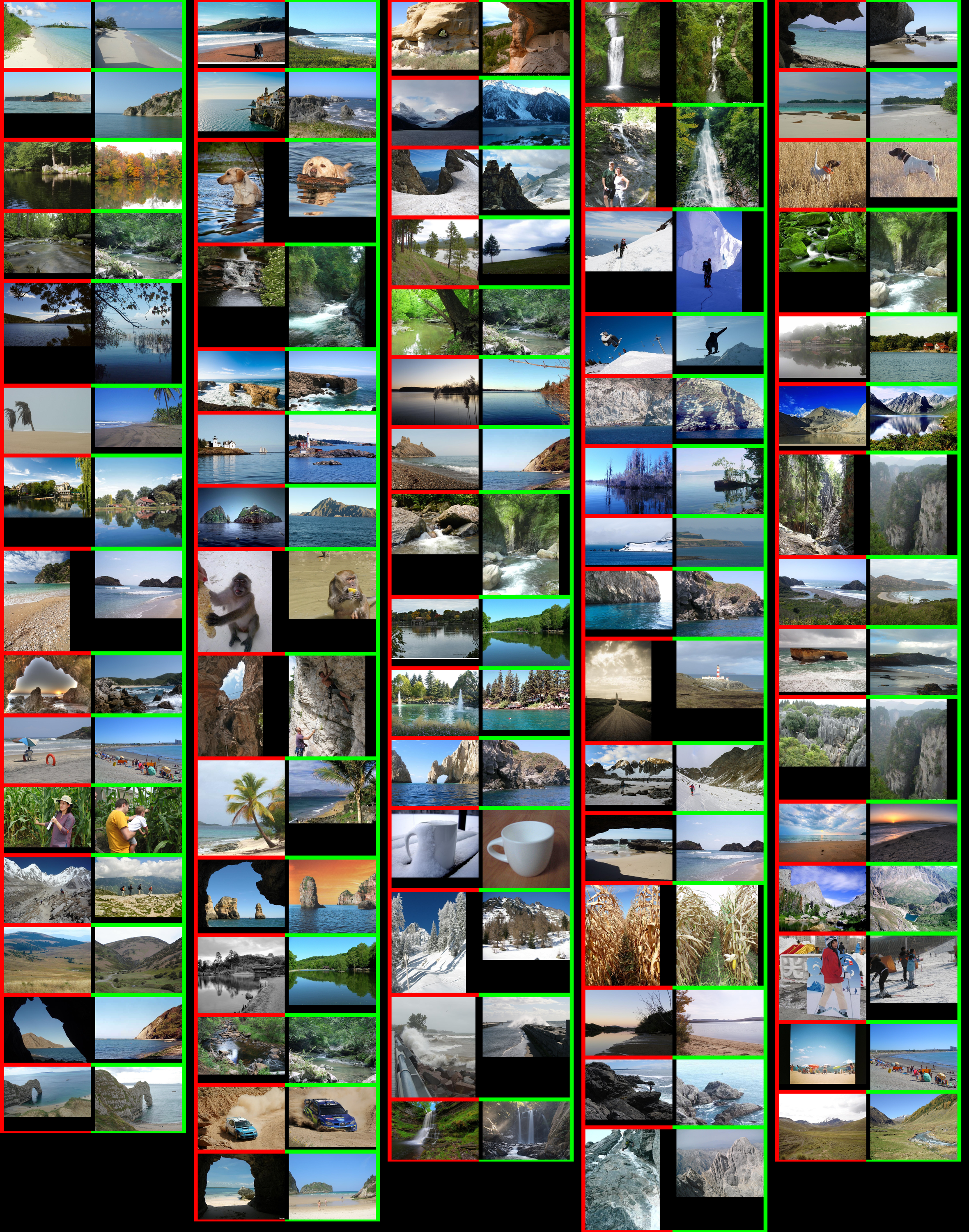}
\Caption[fig:contam-viz-2]{Data Contamination in Places}{Continuation of Figure \ref{fig:contam-viz-1}, showing overlap between Places \cite{zhou2017places} (red bordered images) and ImageNet training data (green bordered images). The overlap between known training data and test data mislabeled as unknowns hinders evaluations, as correctly identifying a known image (which is mislabeled as unknown) will incorrectly penalize an algorithm's score.
}
\end{figure*}

\begin{figure*}[t!]
  \centering
  \includegraphics[width=0.68\textwidth]{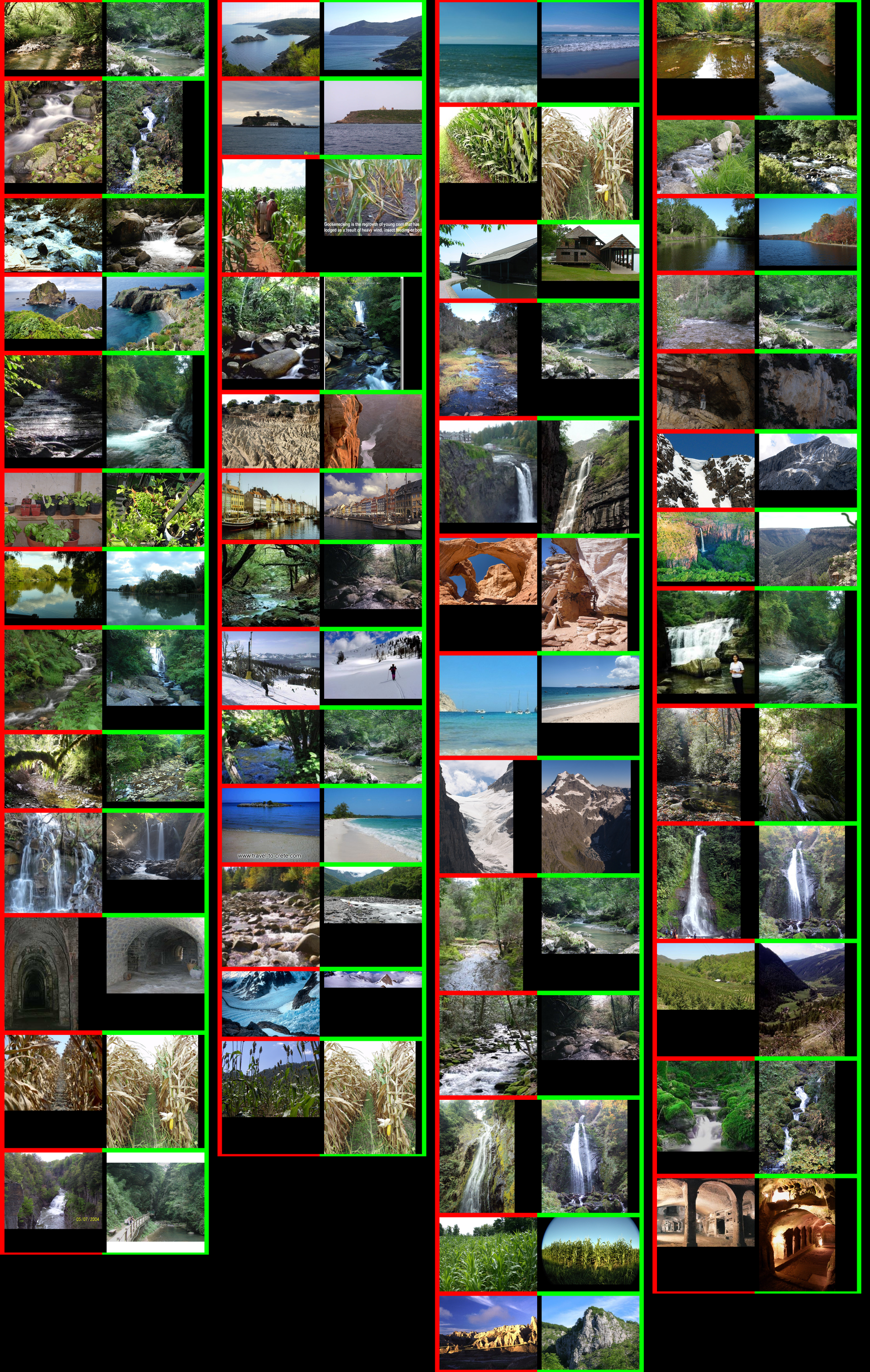}
\Caption[fig:sun-contam-viz-1]{Data Contamination in SUN}{We analyzed images from the OOD dataset SUN \cite{xiao2010sun} (red bordered images) compared with ImageNet training images (green bordered images). To select OOD images from SUN \cite{xiao2010sun}, we examined the validation threshold for PostMax and COSTARR on ViT-H, then selected those images from Places which PostMax would label unknown, but COSTARR would label known. Between this figure and Figure \ref{fig:sun-contam-viz-2}, we present \emph{every image satisfying this constraint}, hence, these images are directly responsible for the performance difference between the algorithms (in terms of OOSA). To select ImageNet training images, we examined all images from the closed-set prediction class (the known class the network predicted the unknown image was) and selected the closest one. As all the images from SUN (red bordered) are treated as unknowns and the network has seen all of the ImageNet training data (green bordered), these falsely labeled unknowns are actually hindering the evaluation, given the clear overlap between known training data and supposed unknowns.
}
\end{figure*}

\begin{figure*}[t!]
  \centering
  \includegraphics[width=0.8\textwidth]{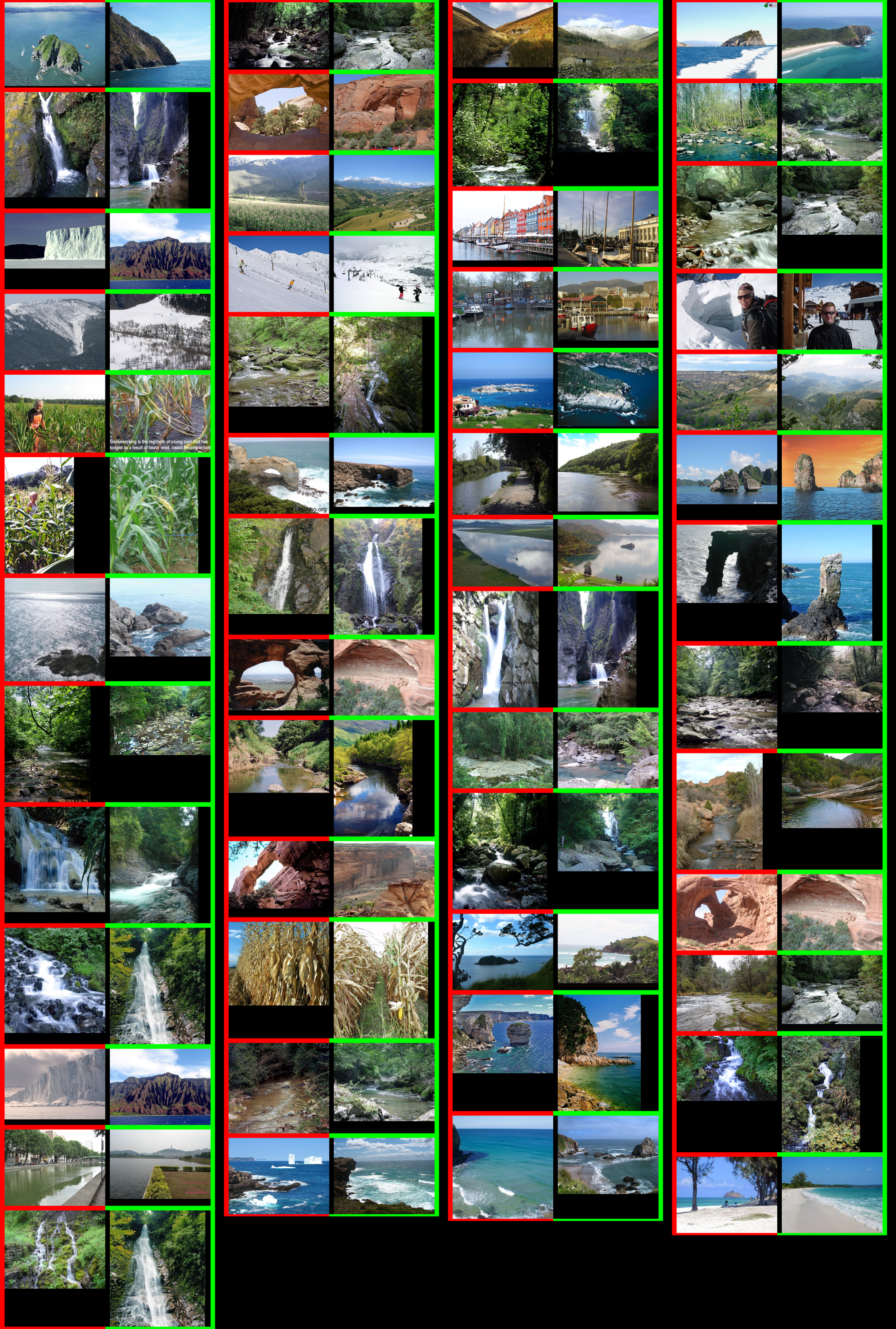}
\Caption[fig:sun-contam-viz-2]{Data Contamination in SUN}{Continuation of Figure \ref{fig:sun-contam-viz-1}, showing overlap between SUN \cite{xiao2010sun} (red bordered images) and ImageNet training data (green bordered images). The overlap between known training data and test data mislabeled as unknowns hinders evaluations, as correctly identifying a known image (which is mislabeled as unknown) will incorrectly penalize an algorithm's score.
}
\end{figure*}

\begin{table}[t]
\begin{tabular}{c?c?cccc}
\multicolumn{1}{c?}{\rotatebox[origin=c]{90}{{\textbf{Arch}}}} & \multicolumn{1}{c?}{\textbf{Method}} & \textbf{Places} & \textbf{SUN} & \textbf{21K E} & \textbf{21K H} \\ \noalign{\hrule height 1.0pt}
\multirow{6}{*}{\rotatebox[origin=c]{90}{{ResNet-50}}} & SCALE & 0.503 & 0.539 & 0.382 & 0.355 \\
 & NNGuide & 0.607 & 0.626 & \textbf{0.776} & 0.487 \\
 & MaxLogit & 0.631 & 0.652 & 0.632 & 0.535 \\
 & MSP & 0.679 & 0.696 & 0.687 & \textbf{0.573} \\
 & PostMax & 0.690 & 0.719 & 0.739 & 0.485 \\
 & COSTARR & \textbf{0.691} & \textbf{0.726} & 0.727 & 0.535 \\ \hline
\multirow{6}{*}{\rotatebox[origin=c]{90}{{ConvNeXt-L}}} & SCALE & 0.588 & 0.617 & 0.514 & 0.492 \\
 & NNGuide & 0.683 & 0.699 & 0.656 & 0.464 \\
 & MaxLogit & 0.652 & 0.675 & 0.586 & 0.535 \\
 & MSP & 0.722 & 0.733 & 0.704 & \textbf{0.620} \\
 & PostMax & \textbf{0.752} & \textbf{0.773} & 0.733 & 0.533 \\
 & COSTARR & 0.739 & 0.767 & \textbf{0.743} & 0.582 \\ \hline
  \multirow{6}{*}{\rotatebox[origin=c]{90}{{ConvNeXtV2-H}}} & SCALE & 0.667 & 0.681 & 0.580 & 0.515 \\
 & NNGuide & 0.720 & 0.727 & 0.720 & 0.552 \\
 & MaxLogit & 0.707 & 0.721 & 0.637 & 0.560 \\
 & MSP & 0.750 & 0.755 & 0.728 & \textbf{0.648} \\
 & PostMax & \textbf{0.779} & \textbf{0.790} & \textbf{0.740} & 0.572 \\
 & COSTARR & 0.730 & 0.748 & 0.712 & 0.565 \\ \hline
\multirow{6}{*}{\rotatebox[origin=c]{90}{{ViT-H}}} & SCALE & 0.637 & 0.672 & 0.569 & 0.510 \\
 & NNGuide & 0.700 & 0.696 & 0.554 & 0.393 \\
 & MaxLogit & 0.692 & 0.717 & 0.632 & 0.566 \\
 & MSP & 0.739 & 0.751 & 0.704 & \textbf{0.627} \\
 & PostMax & \textbf{0.765} & \textbf{0.784} & 0.732 & 0.541 \\
 & COSTARR & 0.754 & 0.781 & \textbf{0.742} & 0.588 \\ \hline
\multirow{6}{*}{\rotatebox[origin=c]{90}{{Hiera-H}}} & SCALE & 0.624 & 0.649 & 0.520 & 0.480 \\
 & NNGuide & 0.749 & 0.756 & 0.677 & 0.485 \\
 & MaxLogit & 0.693 & 0.710 & 0.608 & 0.548 \\
 & MSP & 0.746 & 0.752 & 0.698 & \textbf{0.622} \\
 & PostMax & \textbf{0.773} & 0.785 & 0.727 & 0.548 \\
 & COSTARR & 0.765 & \textbf{0.786} & \textbf{0.743} & 0.612
\end{tabular}
\Caption[tab:oosa-contam]{Operational Open-Set Accuracy}{The mean OOSA ($\uparrow$) of all methods on the contaminated datasets (same experimental setup as \tab{oosa}), which have significant overlap with ImageNet data \cite{bitterwolf2023or}. We additionally validated this overlap for Places \cite{zhou2017places} and SUN \cite{xiao2010sun} in Figures \ref{fig:contam-viz-1}, \ref{fig:contam-viz-2}, \ref{fig:sun-contam-viz-1} and \ref{fig:contam-viz-2}. COSTARR (ours) is, and the best scores for each respective architecture and unknowns dataset are in \textbf{bold}.}
\end{table}

\clearpage

\end{document}